\documentclass{article}
\usepackage{ijcai23}

\usepackage{booktabs}
\usepackage{multirow}
\usepackage{ulem}
\usepackage{times}
\usepackage{array}
\usepackage{soul}
\usepackage{url}
\usepackage{amssymb}
\usepackage[hidelinks]{hyperref}
\usepackage[utf8]{inputenc}
\usepackage[small]{caption}
\usepackage{subcaption}
\usepackage{graphicx}
\usepackage{amsmath}
\usepackage{amsthm}
\usepackage{booktabs}
\usepackage{algorithm}
\usepackage{algpseudocode}

\usepackage[switch]{lineno}

\usepackage{xcolor}
\usepackage{xspace}
\newcommand{\eat}[1]{}

\newcommand{\stitle}[1]{\vspace{0.5ex}{\bf #1}}  % \noindent
\newcommand{\ie}{\emph{i.e.,}\xspace}
\newcommand{\eg}{\emph{e.g.,}\xspace}

\newcommand{\model}{{CIG}\xspace}
\title{Controllable Data Augmentation Training for Image Ordinal Regression}
\title{Robust Image Ordinal Regression with Controllable Data Augmentation}
\title{Robust Image Ordinal Regression with Controllable Image Generation}

\author{
Yi Cheng$^1$\and
Haochao Ying$^2$\thanks{Corresponding Author.}\and
Renjun Hu$^3$\and
Jinhong Wang$^4$\and
WenHao Zheng$^4$\and \\
Xiao Zhang$^5$\and
Danny Chen $^6$\And
Jian Wu$^{2,7}$
\affiliations
$^1$School of Software Technology, Zhejiang Univerisity\\
$^2$School of Public Health, Zhejiang University\\
$^3$Alibaba Group\\
$^4$College of Computer Science and Technology, Zhejiang University\\
$^5$School of Computer Science and Technology, Shandong University\\
$^6$Department of Computer Science and Engineering, University of Notre Dame\\
$^7$Second Affiliated Hospital School of Medicine, Zhejiang University
\emails
\{chengy1, haochaoying, wangjinhong, zhengwenhao, wujian2000\}@zju.edu.com,
renjun0hu@gmail.com,
xiaozhang@sdu.edu.cn,
dchen@nd.edu
}
\begin{document}

\maketitle
\begin{abstract}
Image ordinal regression has been mainly studied along the line of exploiting the order of categories. However, the issues of class imbalance and category overlap that are very common in ordinal regression were largely overlooked. 
As a result, the performance on minority categories is often unsatisfactory. In this paper, we propose a novel framework called \model based on controllable image generation to directly tackle these two issues. Our main idea is to generate extra training samples with specific labels near category boundaries, and the sample generation is biased toward the less-represented categories. To achieve controllable image generation, we seek to separate structural and categorical information of images based on structural similarity, categorical similarity, and reconstruction constraints. We evaluate the effectiveness of our new \model approach in three different image ordinal regression scenarios. The results demonstrate that \model can be flexibly integrated with off-the-shelf image encoders or ordinal regression models to achieve improvement, and further, the improvement is more significant for minority categories.

\end{abstract}

\section{Introduction}

Ordinal classification, which is also widely known as ordinal regression, is a specific type of classification task in which the categories follow a natural or logical order. Category orders are quite common in computer vision tasks, such as human age, image quality, and disease degrees of lesions. Therefore, image ordinal regression has been extensively applied to a number of diverse scenarios, ranging from image quality ranking~\cite{sord} and monocular depth estimation~\cite{KITTI} to clinical image analysis (\eg Gleason grading of prostate cancer~\cite{prostate} and embryo stage classification/grading~\cite{embryo,blasto}). 

% Ordinal regression is a particular classification task, in which each category is not independent and adjacent categories contain developmental relationships.  Different from traditional classification tasks, each category in ordinal regression is not independent, but there is a progressive relationship.

In the literature, image ordinal regression studies have focused on exploiting the order of categories to boost accuracy. Related work can be roughly divided into regression-based, classification-based, and ranking-based methods. 
Regression methods~\cite{regression_face,regression_face2} treat categorical labels as numerical values and directly use loss functions such as mean absolute/square errors to preserve category orders. But these methods may suffer from the non-stationary characteristics of differences between adjacent categories.
%However, regression-based methods face the non-stationary problem, which means the change between categories is not linear. 
%
Classification methods~\cite{sord,poe,mwr} cast ordinal regression as multi-classification and leverage strategies such as soft labeling and relative order maintenance to incorporate the category relationships.
%encodes the label as a one-hot vector and uses loss functions such as cross-entropy to optimize the model, but one-hot encoding and cross-entropy loss will lose the correlation between categories.
%
Ranking methods~\cite{niu2016ordinal,ranking2} replace the original problem with multiple binary classifications and aggregate binary labels to derive ordinal labels heuristically.
% \cheng{Although the ranking method solves the non-stationary problem in principle, it does not necessarily perform better than the classification-based methods in practical applications.}
Overall, classification methods perform better than the other methods.
%However, in recent work, the experimental results show that the classification-based method performs better, so now it is still the mainstream method, and our subsequent content will also be explained based on it.

\begin{figure}[t!]
    \centering
    \includegraphics[scale=0.45]{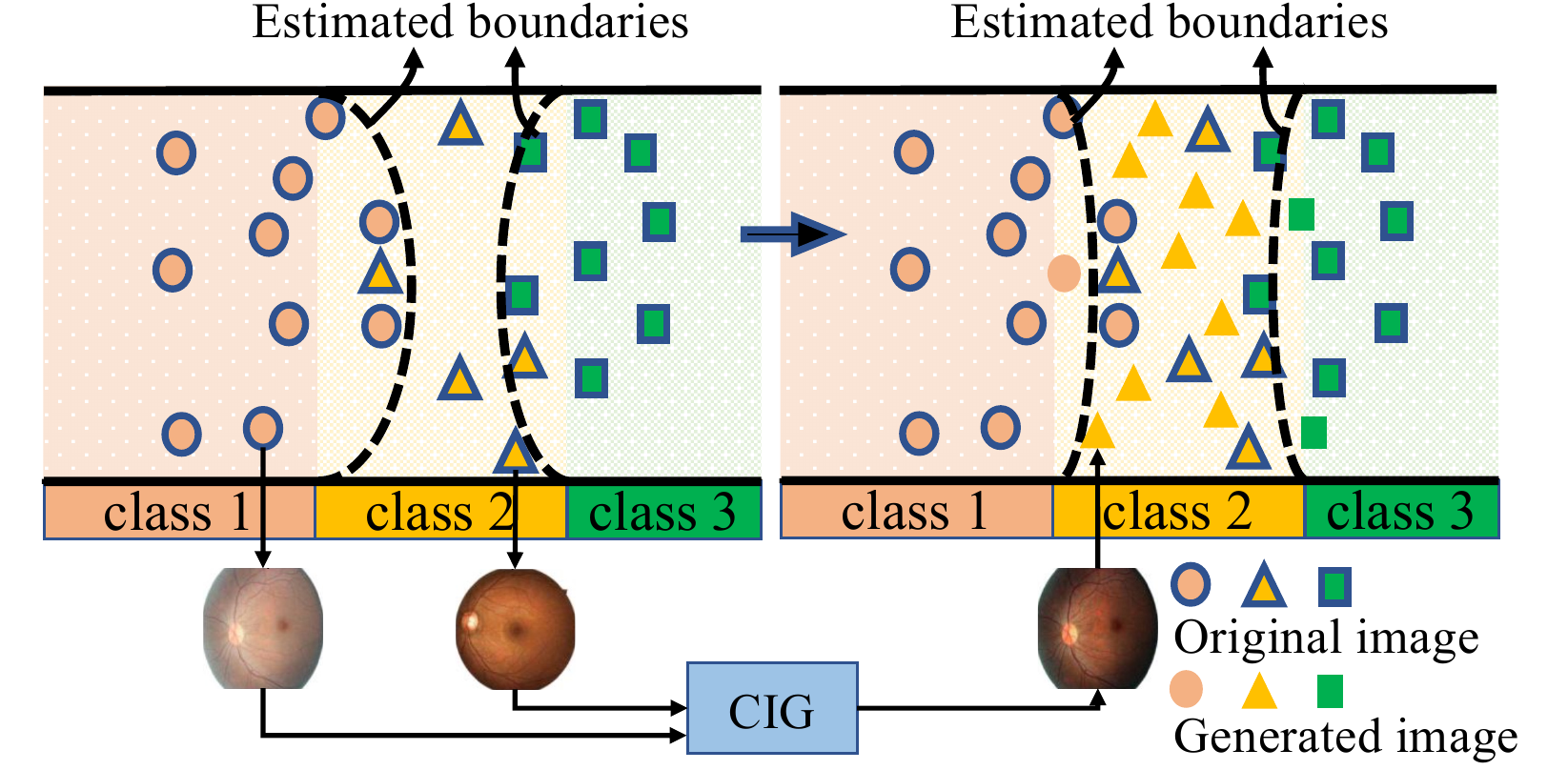}
    %\vspace{-2ex}
    \caption{Illustrating the class imbalance and category overlap issues (left) and controllable image generation (\eg artificial samples of class 2 near the boundaries) to facilitate ordinal regression (right). }
    % An example of image generation: the generated image has the same shape as the left image labeled as 'No DR' but its inner feature is more similar to the right one labeled as 'Mild DR'. Samples of categories 0, 1, and 2 are marked by circles, triangles, and rectangles.
    \label{intro_img}
\end{figure}

While some progress has been made, existing studies largely neglected the issues of \emph{class imbalance} and \emph{category overlap}, which are common in ordinal regression. 
Note that categories in ordinal regression follow certain orders, and it is often the case that data points are not evenly allocated along the metrics associated with these orders. For instance, there is only a very small fraction of images whose quality is rated as excellent or which corresponds to severe organ lesions.
Indeed, we find that the least-sample category in our three datasets only accounts for 5\%, 2\%, and 0.2\% of the total, respectively (Table~\ref{tab:dataset}).
Moreover, categories in ordinal regression are often empirical rather than by definition, \ie they are usually divided by rules. This implies the existence of a certain amount of near- or cross-boundary samples due to feature disturbance and inconsistent subjective judgment. As a result, adjacent categories may overlap (\eg see Fig.~\ref{intro_img}).
The above two issues increase the difficulty for models to learn meaningful category boundaries, especially for the less-represented minority categories. We empirically find that existing approaches~\cite{sord,poe} that ignore these two issues are non-robust, \eg the classification accuracy of minority categories is 
$20 \sim 55\%$ lower than the overall accuracy.  
% $21 \sim 56\%$ lower than the overall accuracy.  
% Adience:60.05~36.08
% DR:79.33~31.21

%the data between the two categories has fuzzy category-related characteristics and will be easily misclassified to adjacent classes.

%However, the previous work mainly focused on the optimization of the neural network and did not propose a feasible solution to the above problems.

In this paper, we propose to tackle image ordinal regression 
%from the perspective of 
by directly addressing the class imbalance and category overlap issues, and develop a novel framework (namely \model) based on controllable image generation.
Our main idea is to generate extra training samples with specific labels near category boundaries, and the generation is biased toward the less-represented categories. 
Fig.~\ref{intro_img} shows an example in which our \model generates extra samples for the minority class 2 near its boundaries. 
As such, each category could be enriched with generated boundary samples, which facilitate learning more accurate and robust decision boundaries. 
Central to our \model approach is the controllable image generation process, \ie producing an artificial image with a specific label near the boundaries. We leverage a separation-fusion-generation pipeline to implement this process. More specifically, we first separate the structural and categorical information of images. Three objectives, \ie structural similarity, categorical similarity, and reconstruction constraints, are introduced to supervise the separation.
As a side effect, the image encoder is also enforced to extract better categorical features for classification.
Afterward, we fuse the structural information of one image with the categorical information of another image to generate the required one.

We conduct extensive experiments on three highly different image ordinal regression scenarios (datasets), \ie age estimation (Adience), diabetic retinopathy diagnosis (DR), and image quality ranking (Aesthetics), to evaluate the effectiveness of our \model approach.
We find that \model can be flexibly integrated with off-the-shelf image encoders (VGG~\cite{vgg} and PVT~\cite{pvt}) or ordinal regression models (POE~\cite{poe}) to attain improvement. \model integrated with the PVT encoder achieves new state-of-the-art classification accuracy and mean absolute error results on all of the three tested datasets. 
Moreover, we empirically show that \model is more friendly to minority categories and the classification accuracy overall and on minority categories  is increased by (1.8\%, 4.5\%), (0.4\%, 5.5\%), and (0.21\%, 8.9\%) on the three datasets, respectively, compared with the best-known baselines.

The main contributions of our work are as follows:
\begin{itemize}
    \item We tackle image ordinal regression by directly addressing the class imbalance and category overlap issues, and present one of the first such methods in the literature.

    \item We propose a new plug-and-play framework \model for the problem. The main novelty behind \model is controllable image generation enabled by separating and fusing structural and categorical information of images.

    \item We verify the effectiveness and improved robustness of \model on three different image ordinal regression tasks.
    
\end{itemize}

\section{Related Work}

In this section, we review related work on image ordinal regression and briefly overview the ideas of generation networks and self-supervised learning that inspire our work.

%\ying{Ordinal regression has gained some momentum thanks to the great success of deep learning frameworks.\old{and methods in the field are almost based on deep neural networks (DNN) in the past years, thanks to the increasing development and improvement of deep learning frameworks.} 
%\old{However, we believe that the previous methods are all aimed at the modification of the model to make the model have a stronger predicting ability, but they do not solve the problem of feature confusion and minority class ignoring in the ordinal regression task. Therefore, we hope to generate images into minor classes and category edges and use these images to strengthen the classification performance of the model.}

\stitle{Image ordinal regression}.
Existing studies %on this topic 
can be classified into regression-, classification-, and ranking-based methods. %\yingcom{give one sentence to conclude the regression-based methods.}
Regression methods treat categorical labels as numerical values and apply optimization. For instance, in \cite{regression_face,regression_face2}, multiple linear regressions were utilized after dimensionality redundancy of the original image space was reduced with subspace learning.
% For instance, \cite{regression_face} proposed the canonical correlation analysis (CCA) methods to extracted age patterns and utilized multiple linear regression for age estimation. \cite{regression_face2} converted the raw image into a low-dimensional representation, then manually adjusted the order of the fitting function and performed regression through different regression methods. 
%
Classification methods cast ordinal regression as multi-classification and emphasize on properly incorporating correlation between categories. SORD~\cite{sord} replaced the traditional one-hot label encoding with soft probability distributions, which allowed models to learn intra-class and inter-class relationships. 
POE~\cite{poe} represented a data point as a multivariate
Gaussian distribution rather than a deterministic point in
the latent space, and exploited the ordinal nature of regression via an ordinal distribution constraint.
MWR~\cite{mwr} leveraged a moving window to refine the prediction of one image based on the supervision of its reference images from adjacent categories.
Ranking methods transform an ordinal regression problem into a series of binary classification sub-problems. 
In \cite{niu2016ordinal}, a multiple output CNN learning algorithm was proposed to collectively solve these sub-problems, and the correlation between these tasks was explored.
A ranking-based ordinal loss was developed to ensure that predictions farther from the true label would incur a larger penalty \cite{ranking2}.
%in order to solve the non-stationary problem,~\cite{niu2016ordinal} first proposed a ranking\ying{-based} method that \ying{further} converts multi-classification problem\old{s} into multiple binary classification problem\old{s}. Later,~\cite{ranking2} migrated the ranking\ying{-based} method to the monocular depth estimation task and achieved \old{good}\ying{better} results.

Different from the previous studies, in this paper, we study image ordinal regression from the perspective of directly addressing the class imbalance and category overlap issues. These two issues are very common in ordinal regression but were largely ignored by previous work. Our work also focuses on the robustness of minority categories which we strongly believe is worth more thorough investigations.

%However, the above methods do not solve the problems that the image at the class edge has blurry features and scarcity of this image. So we hope to generate boundary images through generative networks and enhance the model's ability to extract key features through self-supervision.

\begin{figure*}[t!]
    \centering
    \includegraphics[width=0.9\textwidth]{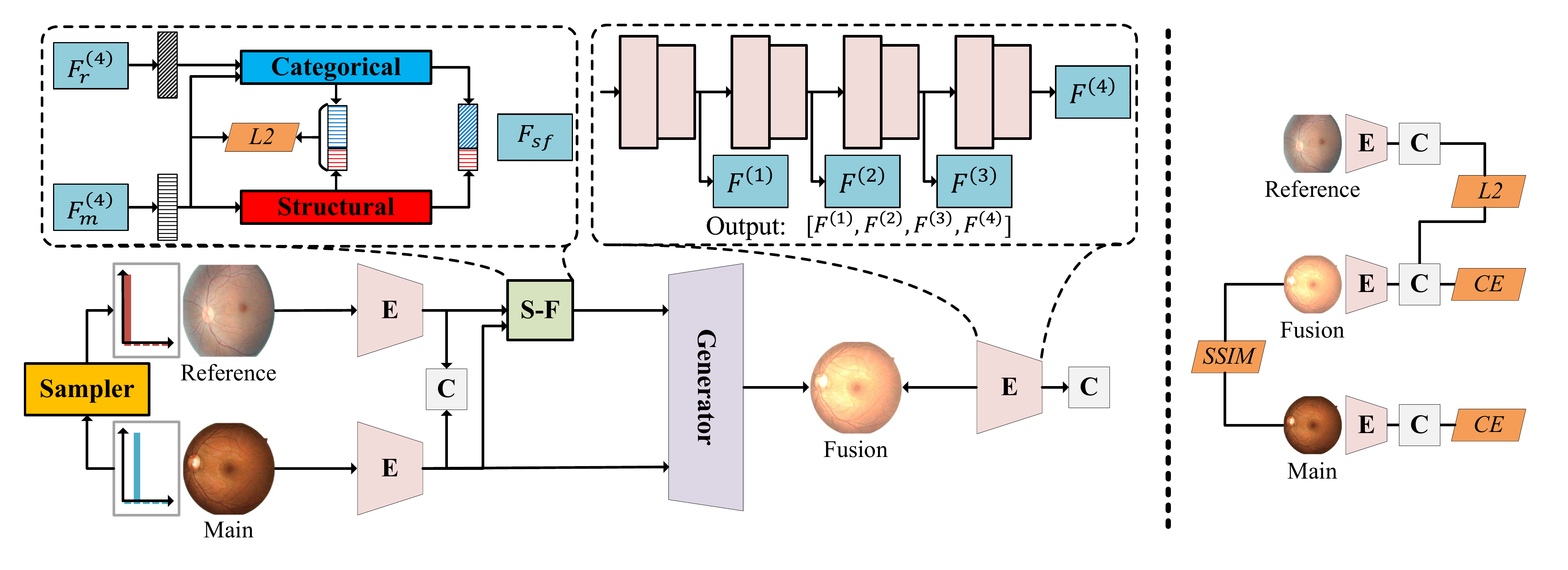}
    
    \caption{An overview of our \model framework. The encoders (E) have the same architecture and weights, and so do the classifiers (C). }
    \label{overview}
\end{figure*}

\stitle{Generation networks} aim to produce images based on feature map vectors. U-Net~\cite{unet} and MAE~\cite{mae} are two representative generation networks which are commonly used in conjunction with popular CNN-based and transformer-based encoders, respectively.
During generation, U-Net connects the encoder and decoder layers that have the same feature map shape. This design makes it possible to contain both high-resolution information and high-dimensional abstract information in the decoder. 
%
%In transformer-based DNNs, because the transformer structure does not change the shape of the feature map, the upsampling module is not required in the decoder. 
MAE was originally proposed as an image auto-encoder, and was trained by recovering an image from its masked version.
%Although the main contribution of MAE is self-supervision, 
We note that the lightweight decoder of MAE is very suitable as a generation network given feature maps produced by transformer-based encoders. 
% MAE is asymmetric whose encoder is much deeper than its decoder, but the lightweight decoder \old{is enough to} plays a key role \old{in MAE} to reconstruct an image similar to the original image. 

Our \model framework adopts a generation network to produce extra training samples. Differently, we fuse two images to generate the artificial one and require the image generation process to be controllable. %, \ie generating an image with a specific label near a category boundary. 
This is implemented by a separation-fusion-generation pipeline.

%\With the remarkable feature extraction capacity, CNN and transformer have become the mainstream encoder architecture. Thus, in order to meet the feature map structure of \ying{these two} different encoders, 

%\old{After the image is generated by the generative network,} \ying{Through designing a generative network, we hope that this self-supervised framework will strengthen the ability of the classification network to further extract key distinguished features between adjacent categories}. At the same time, the generator also \old{needs}\ying{contributes} to obtain better additional data \old{for data augmentation under self-supervision}.

\stitle{Self-supervision learning} can obtain representations to help downstream tasks by learning from some auxiliary tasks. In~\cite{orthogonal}, the authors proposed to partition an image into private and invariant domains via a domain-separation network. The separation network was trained based on the self-supervised orthogonality and similarity loss.
%and then restrained the uncorrelation of the decomposed features through an orthogonal loss function. 
%
CycleGAN~\cite{cycleGAN} implemented unpaired image-to-image translation by using a cycle loss that constrained the content consistency between the original and corresponding generated images.
%a loss function is used on the generated image to require the image to be consistent with the original image. 
%
Inspired by these methods, our CIG learns to separate structural and categorical information of images by constraining the structural and categorical similarities between the original and generated images. %and further enforcing the separation reconstruction.
%construct a factorization-reconstructor and constrain the complementarity of feature decomposition and consistency after reconstruction through loss functions.

\section{Methodology} 
\label{sec:method}

%\subsection{Overview of Proposed Framework}

% Most of the current research is to modify the network construction and processing steps. 
In this section, we present our \model framework with controllable image generation. A framework overview is given in Fig.~\ref{overview}. In a nutshell, \model generates extra training samples with specific class labels near category boundaries, to facilitate learning of the task. More specifically, to classify an image, referred to as a main image, \model first samples a reference image from its adjacent categories. Both the main and reference images are passed through the image encoder to extract feature maps. These feature maps are then combined to generate a fusion image through a separation-fusion-generation pipeline, such that the label of the fusion image is the same as the reference image. We exploit both self-supervision and classification supervision to train our \model.

We first introduce the architecture of \model in Section~\ref{subsec:arch}, and then discuss model supervision in Section~\ref{subsec:loss}.

%consisting of a reference image sampler\yingcom{sampler is not shown in Fig.\ref{overview}}, encoder, classifier, and generator. First, \model will sample a non-repetitive image from the training set as the main image, and then the sampler samples a reference image from the adjacent category of the main image. The encoder extract\ying{s} feature maps from these two images and the classifier classif\ying{ies} them. At the same time, these two feature maps will be input into the Combine module for feature-shape fusion. The generator block generates a fusion image (the bottom picture in Fig.\ref{overview}) from the fused feature. Finally, the fusion image will be input into the encoder for classification training. \yingcom{why we design each module should be explained in detail.}

% \model extracts features through the encoder, the classifier performs classification, and the generator fuses the features and generates a characteristic-limited image. During the process of the generator generating images, through the limitation of the loss function, the feature extraction ability of the encoder will be optimized. The generated images also continue to be used for training, which expands the training dataset.

\subsection{Architecture}\label{subsec:arch}

\stitle{Sampler}.
Without loss of generality, we assume that the categories of our target image ordinal regression task are labeled consecutively with integers $1, 2, \dots, K$, where $K$ is the total number of categories.
Given a main image $X_m$ with a label $m$, our \model first samples a reference image $X_r$ (with a label $r$) from the training set such that $|m-r|=1$, \ie $X_m$ and $X_r$ are from adjacent categories. \model then generates an artificial fusion image $X_f$ of label $r$ based on $X_m$ and $X_r$. This adjacent sampling assures that the fusion image is near the boundary between the categories $m$ and $r$.

%Since non-adjacent images may generate images outside of these two categories, which makes the self-supervised loss function less accurate.

We allow the generation process to bias toward minority categories that are originally less represented. We consider two samplers for \model. 
The first one is an equal sampler, which samples $X_r$ from adjacent categories with equal probability. 
The second one is called an inverse-ratio sampler. 
Let  $N_{m-1}$ and $N_{m+1}$ denote the numbers of raw images in categories $m-1$ and $m+1$.
The probabilities of sampling a reference image from categories $m-1$ and $m+1$ are then determined by ${N_{m+1}}/N_{adj}$ and ${N_{m-1}}/N_{adj}$, respectively, where $N_{adj}={N_{m-1} + N_{m+1}}$.
Both samplers can increase the proportion of training samples, \ie the main %\yingcom{increase the samples of main or reference images?} 
and fusion images, for minority categories. 

%if consider real and artificial images together.

%\subsection{Encoder and Classifier}

\stitle{Image encoder and classifier}.
The main and reference images are then passed through an image encoder to extract feature maps for subsequent generation and classification. Technically, \model can use all image encoder architectures, and we explore the classic CNN-based architectures (\eg VGG16~\cite{vgg}) and the more recent Pyramid Vision Transformer (PVT)~\cite{pvt} architecture as the encoder in this work. 
A single fully-connected (FC) layer is adopted as the classification head which takes the extracted feature maps of images as input and outputs the predictive probabilities of all the categories.

% In \model, the single Fully-Connected-Layer (FC) is used as the , and the label is encoded to the one-hot vector. 

%Because the transformer architecture is outstanding in computer vision tasks recently, we use and the transformer-based architecture as encoders in \model: and.

\stitle{Controllable generator}.
A core of \model is controllable image generation through which we produce a fusion image $X_f$ with a label $r$ using the main and reference images $X_m$ and $X_r$. We implement the generation process using a separation-fusion-generation pipeline, and propose to separate the structural and categorical information of images.\footnote{The concept of structural information is borrowed from structural similarity (\url{https://en.wikipedia.org/wiki/Structural\_similarity}).}
Structural and categorical information is the information for determining the overall structure and the category of an image, respectively.

The upper-left part of Fig.~\ref{overview} illustrates the separation-fusion {(S-F)} operation. Specifically, we use either two $1\times1$ convolutional layers (for a CNN-based encoder) or two FC layers (for a Transformer-based encoder) as the structural and categorical extractors to extract relevant features from feature maps $F_{m}^{(4)}$ and $F_{r}^{(4)}$. The two extracted feature maps are concatenated into $F_{sf}$. 
The channel numbers (or lengths) of the feature maps from the structural and categorical extractors are set as $\tau$ or $1-\tau$ times the channel number (or length) of the last feature map $F^{(4)}$ from the encoder, respectively, where $\tau$ is a percentage value for controlling the structural information proportion of the concatenated feature map from the S-F module.
% The dimensionality of the vectors from the structural and categorical extractors is set to $\tau$ and $1-\tau$ of the one of feature maps, respectively.
%
The computed structural feature map of $X_m$ and categorical \eat{vector} feature map of $X_r$ are concatenated to generate the fusion image $X_f$. As such, $X_f$ is desired to be of class $r$ while being structurally similar to $X_m$, \ie lying near the boundary between categories $m$ and $r$. Further, the categorical \eat{vector} feature map of $X_m$ is extracted and concatenated with its structural \eat{counterpart} feature map for regularization (we will explain this below). 
\model adopts UNet or the decoder of MAE as the generation network, depending on the encoder architecture (\ie CNN-based or Transformer-based).
%
% \replaced{In CNN-based \model, we replaced the encoder part of UNet with Vgg16, and the skip-connection in UNet is only for $X_m$'s feature maps. In transformer-based \model, inspired by MAE, a single-layer lightweight transformer layer is used as a generator. Compared with ViT~\cite{vit} used in MAE, PVT has a pyramid feature structure, that is, the output features of the last layer will become smaller, so we restore the feature shape through the repeat operation.}{I could not understand this part, please rephrase these sentences to describe details/difference of the two generation networks. Also explain how the feature map of $X_m$ and the concatenated vector are used by the generator.}
% 
% \cheng{In the \model(Vgg) model, the concatenated vector and the four feature maps of $X_m$, $[f_1,f_2,f_3,f_4]$ in Fig.~\ref{overview}, are input uniformly into UNet for generating $X_f$. But in the \model(PVT) structure, we only use the concatenated vector as the input of the MAE-decoder. In order to ensure that the size of the generated image is consistent with the original image, the concatenated vector is copied 4 times by the repeat operation.}

\eat{
\begin{figure}[t]
    \centering
    \includegraphics[scale=0.6]{img/SFR.pdf}
    \caption{Combine Module. The green rectangle indicates the feature map of the main image while the blue one is for the reference image. Concatenation operation is used to combine the main image's shape information and the reference image's feature information.}
    \label{encoder}
\end{figure}
}

\begin{figure}[t]
    \centering
    \includegraphics[scale=0.30]{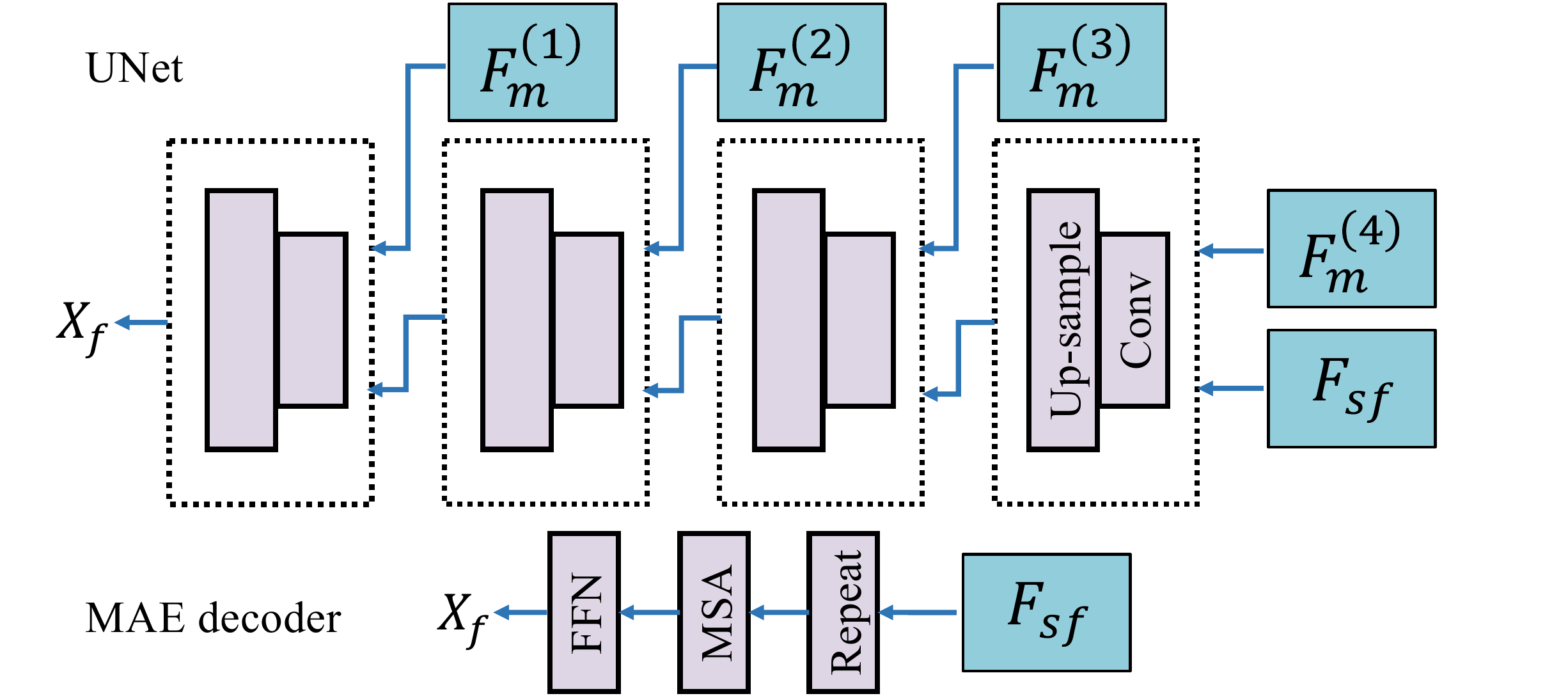}
    \caption{Two types of generation networks in \model.}
    % \marked{$f_{s-f}$ to $F_{sf}$, $f_{i-main}$ to $F_m^{(i)}$ in both Fig. 2 and 3}} 
    % $f_{s-f}$ is the feature map output from separation-fusion-generation pipeline. Transposed 2D convolutional layer is used as up-sampling layer. FFN and MSA represent feed-forward network and multi-head self attention in transformer architecture.
    \label{fig:generator}
\end{figure}

As shown in Fig.~\ref{fig:generator}, in UNet, the concatenated 
%\eat{vector} 
feature map $F_{sf}$ and the last feature map $F_m^{(4)}$ of $X_m$ are first input into an up-sampling block. This up-sampling process is repeated 
%\eat{multiple} 
three times to generate $X_f$. In the repetitions, the concatenated feature map is replaced by the output of the previous block. 
% except when replacing the concatenated vector with the output of the previous block until $X_f$ is generated. 
In the light-weight MAE decoder, a single Transformer block with a {multi-head self-attention} (MSA) layer and a feed-forward network (FFN) is used for the generation, and $F_{sf}$ is repeated four times to ensure size consistency.

We apply three types of self-supervision to ensure the separation-fusion-generation pipeline to behave as we desire.

% In \model, the decoder restores an image with $X_m$'s structure and $X_r$'s features through the fused feature from Combine module. 
% In order to meet the network structure of CNN-based and transformer-based architectures, we use two different decoders: UNet-based and MAE-based Generators. 

%In order to generate a fusion image $X_f$ with both the shape of the main image and the characteristic of the reference image, we design the Combine Module to extract these features and fuse them. 

%As shown in the Fig.\ref{encoder}, the feature maps of the $X_m$ and the $X_r$ are input into the Combine module at the same time, the shape information in the $X_m$ is extracted by the shape extractor, and the feature information of the $X_r$ is extracted by the feature extractor. 
%The proportion of feature information and shape information in the fused feature is the same. Finally, the fusion feature map is obtained through the concatenation operation. The shape or feature extractor in CNN-based \model are both FC layers and these in PVT-based \model are both $1\times1$ convolutional layers. 

%In the process, we propose the reconstruction loss, which will be introduced in section \ref{sec:loss}.

\subsection{Model Supervision}\label{subsec:loss}

\model exploits self-supervision to guide information separation and image generation, and exploits classification supervision to improve the overall performance.

%In our \model, we use loss functions to limit fusion images' information composition and to improve classification performance. Therefore, the overall loss function is made up of two parts, one is the image generation loss, and the other is the prediction loss.

%\subsubsection{Self-Supervision} 

%We hope to generate an image $X_f$ through the generator, which has a similar pattern with $X_m$ but has a similar label with $X_r$. 

\stitle{Self-supervision}.
To ensure the structural extractor works effectively, we require the fusion image $X_f$ to be more structurally similar to the main image $X_m$ than $X_r$. We apply the commonly-used structural similarity index measure (SSIM) to quantify the perception-based similarity between two images $X$ and $Y$, as:
\begin{equation}\label{eq:ssim}
\mbox{SSIM}(X,Y)=\frac{(2\mu_X\mu_Y+c_1)(2\sigma_{XY}+c_2)}{(\mu_X^2\mu_Y^2+c_1)(\sigma_X^2+\sigma_Y^2+c_2)},
\end{equation}
where $\mu_X$ and $\sigma_X^2$ are the pixel sample mean and variance of image $X$, $\sigma_{XY}$ is the covariance of images $X$ and $Y$, and $c_1=(0.01L)^2$ and $c_2=(0.03L)^2$ are variables for stabilizing the division operation 
%with a weak denominator. 
($L$ is a dynamic range of the pixel-values of the images). 
The SSIM values are within $(0, 1]$, with larger values for higher similarity. We then minimize the following structural generation loss in order to enforce structural extraction, as:
\begin{equation}
    \mathcal{L}_{SG}=-\frac{1}{2} \big( \log \mbox{SSIM}(X_m,X_f) + \log(1-\mbox{SSIM}(X_r,X_f)) \big).
    \label{loss:lsg}
\end{equation}

On the other hand, the fusion image $X_f$ should be more categorically similar to the reference image $X_r$ than $X_m$. Thus, we require the predicted categorical probability vectors $P_{f}$ and $ P_{r}$ of $X_f$ and $X_r$ to be similar. Specifically, we minimize the squared Euclidean distance between the two un-normalized probability vectors to enforce categorical extraction. The corresponding categorical generation loss is:
\begin{equation}
    \mathcal{L}_{CG}=||P_{r}-P_{f}||^2,
    \label{loss:lcg}
\end{equation}
where $P_{r} \in \mathbb{R}^K$ and  $P_{f} \in \mathbb{R}^K$ are the raw probability vectors of $X_r$ and $X_f$ output by the classifier.

%have the same label as $X_r$. However, there is no perfect discriminator to judge the label of the generated image.  The feature map output by the encoder contains very rich semantic information, which reflects the characteristics of the image. But it also contains information such as shape, so we use the probability vector after the classifier for loss calculation.

%Therefore, we use Euclidean Distance to measure the distance between feature vectors of $X_r$ and $X_f$. Let the feature vector of $X_m$, $X_r$ and $X_f$ are $F_{cm}$, $F_{cr}$, $F_{cf}$:
% \begin{equation}
    % \mathcal{L}_{GL}=\frac{(F_{cf}-F_{cr})^2}{\sum((F_{cf}-F_{cr})^2)}.
% \end{equation}

Moreover, we desire to use ``simple'' structural and categorical extractors (the simpler the better). In other words, we hope that the encoder can learn to extract categorical information, instead of relying heavily on the extractors, as the categorical information will benefit the subsequent classification.
Therefore, we further optimize a reconstruction loss between the feature map $F_m^{(4)}$ and the concatenated vector $F_{sf}=\mbox{concat}[h_c(F_m^{(4)}), h_s(F_m^{(4)})]$ of the main image, as:
\begin{equation}
    \mathcal{L}_{RC}=||F_m^{(4)} - \mbox{concat}[h_c(F_m^{(4)}), h_s(F_m^{(4)})]||^2,
    \label{loss:lrc}
\end{equation}
where $h_c$ and $h_s$ stand for the categorical and structural extractors of \model.

%Assuming that the Feature block in Fig.\ref{encoder} is $B_f$ and the shape block is $B_s$, and the feature map from the encoder for image $X_m$ is $F_m$, then the reconstruction loss function of this part is:

% shape features and feature features in the Combine module are complementary. 
% In this way, the loss of information can be reduced, and the generated image can be considered as an extension of the data set. 

Finally, the overall self-supervised generation loss is computed as a weighted sum of the above three losses:
\begin{equation}
    \mathcal{L}_{G}=\alpha \cdot \mathcal{L}_{SG} + \beta \cdot \mathcal{L}_{CG} + \mathcal{L}_{RC}.
    \label{eq:loss-generation}
\end{equation}

%\subsubsection{Classification Supervision}
\stitle{Classification supervision}.
We use the traditional cross-entropy (CE) loss to optimize the classification capacity of \model. The CE loss is evaluated and optimized only on the main and fusion images with labels $m$ and $r$, respectively. Let $P_m \in \mathbb{R}^K$ and $P_f \in \mathbb{R}^K$ be the predicted categorical probability vectors of the main and fusion images \eat{$X_m$ and $X_f$} outputted from the classifier. With some abuse of notation, the CE loss on $X_m$ \eat{with label equals to $L_m$} (with label $m$) can be expressed as:
% on $X_m$ can be expressed as:
\begin{equation}
    \mathcal{L}_{CE}(P_m, m)=-\log\frac{\exp(P_m^{m})}
    {\sum_{k=1}^{K} \exp(P_m^k))},
    \label{CEL}
\end{equation}
where $P_m^h$ denotes the value of the $h$-{th} entry in $P_m$. 
% \begin{equation}
%     \mathcal{L}_{CE}(P_m, m)=-\log\frac{\exp(P_{m,m})}
%     {\sum_{k=1}^{K} \exp(P_{m,k}))}.
%     \label{CEL}
% \end{equation}
The overall classification loss of CIG is a weighted sum of cross-entropy on the main and fusion images, as:
% \begin{equation}
%     \mathcal{L}_{C}=\mathcal{L}_{CE}(P_m, m) + \lambda \cdot \mathcal{L}_{CE}(P_f, r).
%     \label{eq:loss-classification}
% \end{equation}
\begin{equation}
    \mathcal{L}_{C}=\mathcal{L}_{CE}(P_m, m) + \lambda \cdot \mathcal{L}_{CE}(P_f, r).
    \label{eq:loss-classification}
\end{equation}

The complete training process of \model is summarized as follows. We first train the entire model (\ie the encoder, classifier, S-F module, and generator) 
in each batch. The encoder and classifier are supervised by the classification loss in Eq.~(\ref{eq:loss-classification}) while the S-F module and generator are optimized by the self-supervision loss in Eq.~(\ref{eq:loss-generation}). 
% Why separately train
We find that the generator-related parameters are optimized much more  slowly than the other parameters, and hence adopt two optimizers with different learning rates for optimization. 
After the generation network has been adequately optimized, we continue to train the encoder and classifier alone for another 8,000 to 24,000 batches. Note that with controllable image generation, the encoder is required to also extract structural information, while this information is not used for classification. The continued training allows the model to better focus on category-related features to further boost accuracy.

%Therefore, we continue train the encoder and classifier after the whole \model is trained completely to increase the proportion of categorical-relative evidence and remove the unnecessary component. 

%On the three datasets, the continue training starts after 8000 iterations (on the Adience and Aesthetics datasets) or 24000 iterations (on the DR datasets).

\section{Experiments} \label{sec:exp}

\begin{table}[t]
    \centering
    \begin{tabular}{c c c}
        \hline
        Dataset     & \# of images & \% of images in each category \\
        \hline
        Adience     & 17,321    & 14 / 12 / 12 / 10 / 29 / 13 / 5 / 5 \\
        DR       & 35,126    & 74 / 7 / 15 / 3 / 2	\\
        Aesthetics     & 13,706    & 2 / 24 / 66 / 8 / 0.2	\\
        \hline
    \end{tabular}
    \caption{Dataset statistics. The categories are arranged in order.}
    \label{tab:dataset}
\end{table}

We evaluate the effectiveness of our \model approach on three different image ordinal regression tasks. 
Four sets of experiments are conducted to evaluate: (i) the overall effectiveness of \model compared with known state-of-the-art methods, (ii) the robustness for minority categories, (iii) the contribution of each component in \model, and (iv) the parameter sensitivity.

\subsection{Experimental Setup} \label{subsec:exp-setup}

\stitle{Datasets.} We use three public datasets to evaluate our CIG. 

%\noindent
(1) {\bf Adience}~\cite{adience} is a face image dataset from Flickr. Its categories correspond to human ages. 

%\noindent
(2) {\bf DR} (Diabetic Retinopathy)~\cite{dr} contains high-resolution fundus images of  % 17,563 
patients.\footnote{\url{https://www.kaggle.com/c/diabetic-retinopathy-detection}} 
These images are classified according to the degrees of retina lesions.

% In order to facilitate the diagnosis of diabetic retinopathy, the images are divided into five levels based on damage to the retina, retinal hemorrhages, obvious intraretinal microvascular abnormalities, etc.

%\noindent
(3) {\bf Aesthetics}~\cite{aesthetics} is another Flickr image dataset whose images are rated by the image quality.

%provides 15687 Flickr images URLs but 13706 of them are available now because of copyright etc. 
%In order to reduce the influence of invalid images, we divide the dataset into five-fold with 80\% training images and 20\% validation images, 
% and the results presented in Table~\ref{tab:previous compare} are reproduced average values on our data splitting.

\begin{figure}[t!]
    \centering
    \vspace{-1.5ex}
    \includegraphics[scale=0.2]{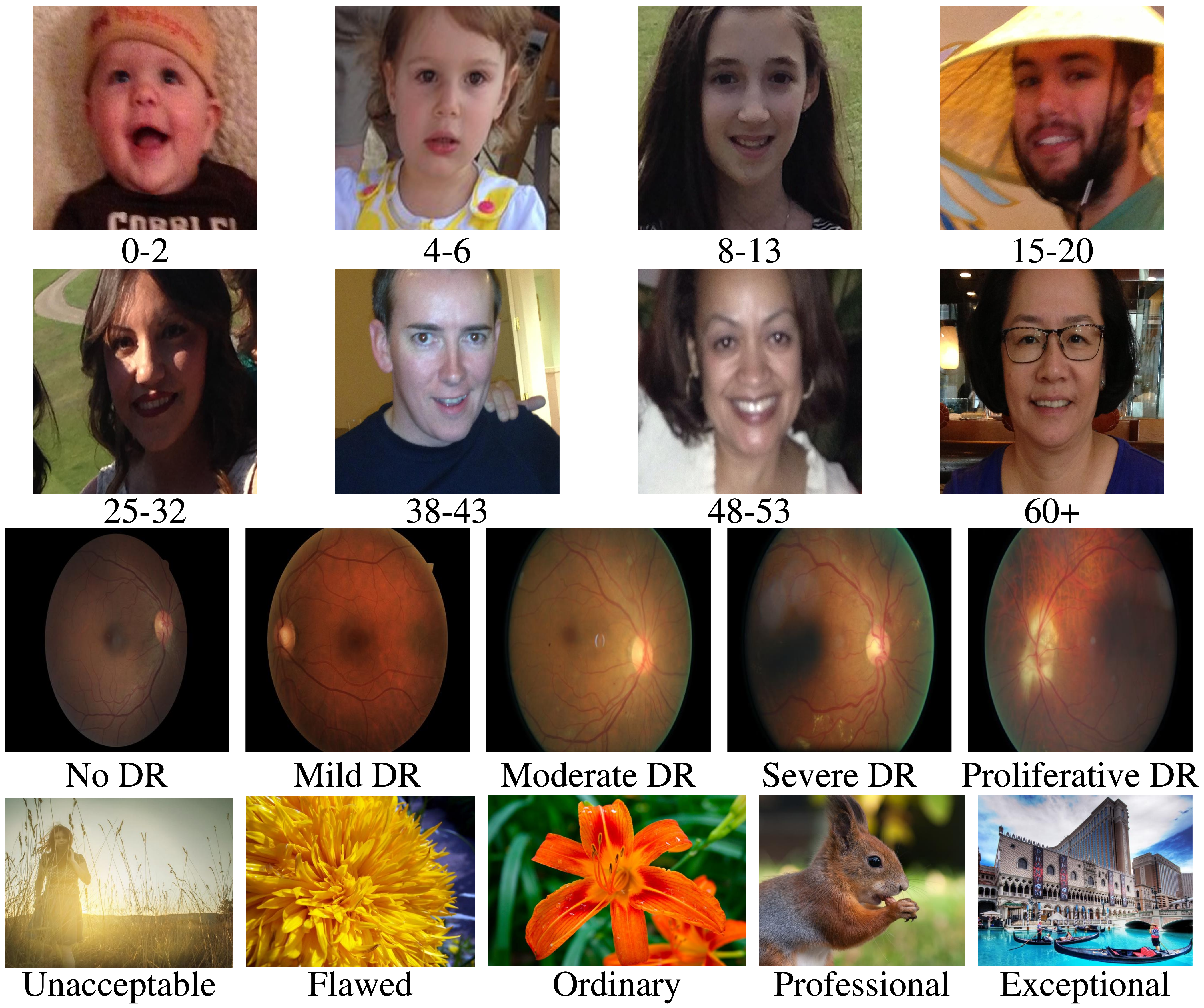}
    \caption{Ordinal categories and example images of the Adience, DR, and Aesthetics datasets (from top to bottom).}
    \label{fig:samples}
\end{figure}

Table~\ref{tab:dataset} summarizes some statistics of these datasets, and Fig.~\ref{fig:samples} illustrates their ordinal categories with example images. Note that image ordinal regression on the three datasets corresponds to
human age estimation, diabetic retinopathy diagnosis, and image quality ranking, respectively.

% including two concrete datasets and one abstract dataset.

\textbf{Metrics.} We adopt classification accuracy (ACC) and mean average error (MAE) between predicted and ground-truth category probabilities for performance evaluation.

\stitle{Implementation.}
Our \model is implemented using PyTorch~\cite{pytorch}, which is available at GitHub\footnote{\url{https://github.com/Ch3ngY1/Controllable-Image-Generation}}. 
The inverse-ratio sampler is used by default and image encoders are initialized with the weights pre-trained on ImageNet1K~\cite{imagenet}. We adopt the default Adam optimizer and a batch size of 18 for model training. The learning rates for the encoder and generator are set as $1\times10^{-4}$ and $5\times10^{-3}$, respectively. 
We optimize hyper-parameters on Adience with $\alpha\in\{1,2,5\}$, $\beta\in\{1,2,5\}$, $\lambda\in\{0, 0.1, \dots, 1\}$, and $\tau \in \{0.1, 0.2, \dots, 0.9\}$, and choose $\alpha=5$, $\beta=2$, $\lambda=0.2$, and $\tau=0.2$ for all our tests.  
We use 5-fold (on Adience and Aesthetics) or 10-fold (on DR) cross-validation, and report the average results.
% in the following. 
All the experiments are conducted on a machine with 16 Intel(R) Xeon(R) Gold 6226R 2.90GHz CPUs and an NVIDIA RTX 3090 GPU.
% The source code 

\begin{table*}[ht]
    \centering
    \label{tab:summary}
    \begin{tabular}{cccccccc}
        % \toprule
        \hline
        \multirow{2}{*}{Method} & \multicolumn{2}{c}{Adience} & \multicolumn{2}{c}{DR} & \multicolumn{2}{c}{Aesthetics} \\
        % \cmidrule{2-7}
        % \cline{2-7}
        & ACC (\%)$\uparrow$ & MAE$\downarrow$ & ACC (\%)$\uparrow$ & MAE$\downarrow$ & ACC (\%)$\uparrow$ & MAE$\downarrow$ \\
        % \midrule
        \hline
        CNNPOR~\cite{cnnpor} & 57.4 & 0.55 & 82.87 & 0.335 & 67.48 & 0.354  \\
        GP-DNNOR~\cite{gpdnnor} & 57.4 & 0.54 & -- & -- & -- & -- \\
        MT~\cite{MT} & -- & -- & 82.80 & 0.360 & -- & --\\
        Poisson~\cite{Poisson} & -- & -- & 77.10 & 0.380 & -- & --\\
        SORD~\cite{sord} & 61.03\eat{59.6} & 1.49\eat{0.49} & 78.67  & 1.421 & \uline{69.97} & 0.567\\
        POE~\cite{poe} & 59.3\eat{60.5} & 0.49\eat{0.47} & 80.48 & \uline{0.312} & 68.92 & \uline{0.351}\\
        MWR~\cite{mwr} & \uline{62.6} & \uline{0.45} & -- & -- & -- & --\\
        % \hline
        \model-VGG (ours) & 61.4 & 0.47 & \uline{82.94} & 0.326 & 67.48 & 0.387\\
        \model-PVT (ours) & \textbf{64.4} & \textbf{0.43} & \textbf{83.27} & \textbf{0.304}& \textbf{70.18} & \textbf{0.343}\\
        \hline
        % \bottomrule
    \end{tabular}
    \caption{Performance comparison of CIG and known methods. The best and second-best results are marked in {\bf bold} and underlined, respectively.  `--' indicates that we cannot find or reproduce the results due to private implementation of the original papers or inapplicable settings.}
    \label{tab:previous compare}
\end{table*}

\begin{figure*}[t!]
    \centering
    \hspace{-3ex}
    \begin{subfigure}[b]{0.27\textwidth}
         \centering
         \includegraphics[width=\textwidth]{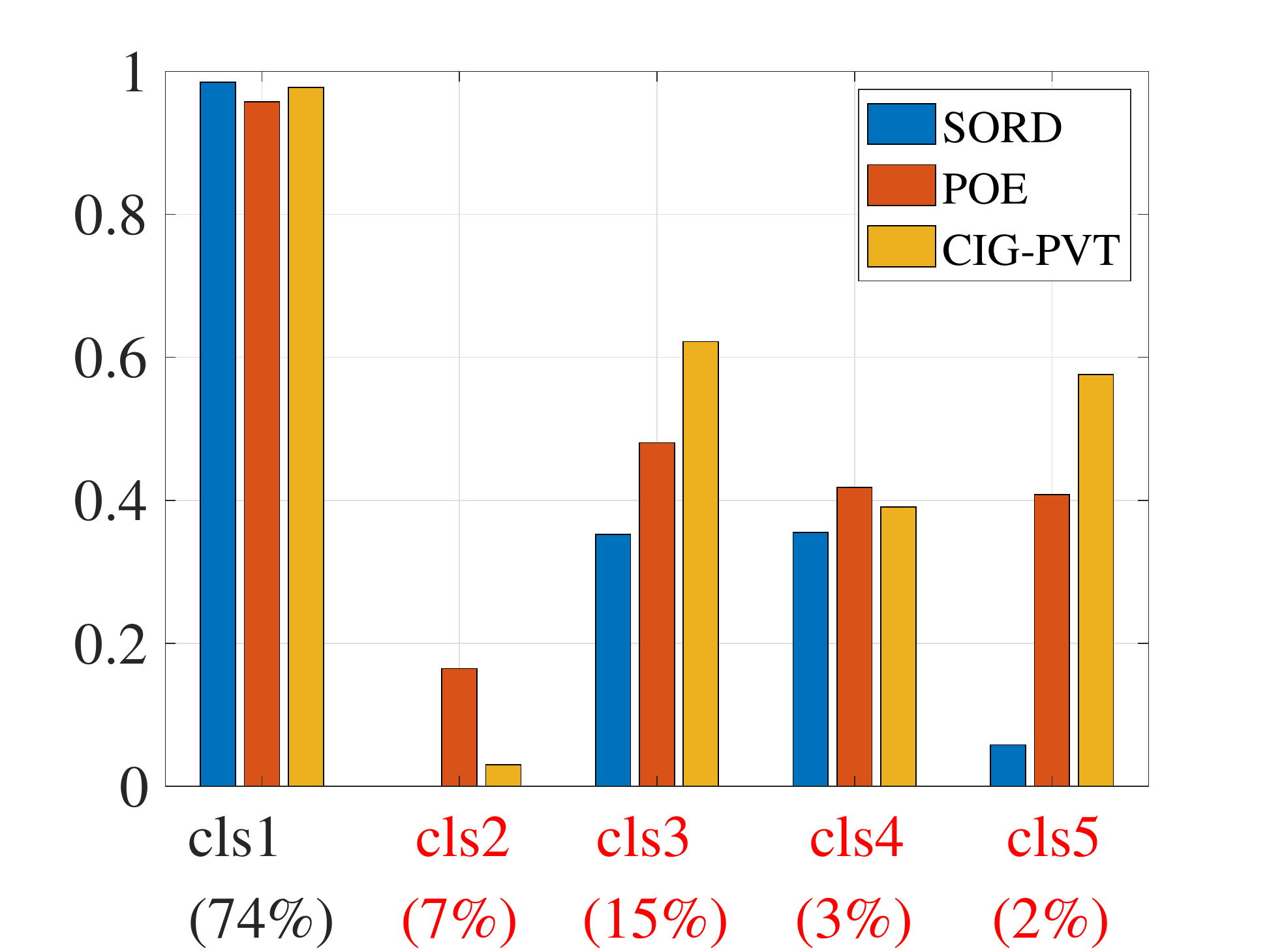}
         \caption{ACC on DR}
         \label{fig:robust_acc_dr}
     \end{subfigure}
     %\hfill
     \hspace{-3.75ex}
     \begin{subfigure}[b]{0.27\textwidth}
         \centering
         \includegraphics[width=\textwidth]{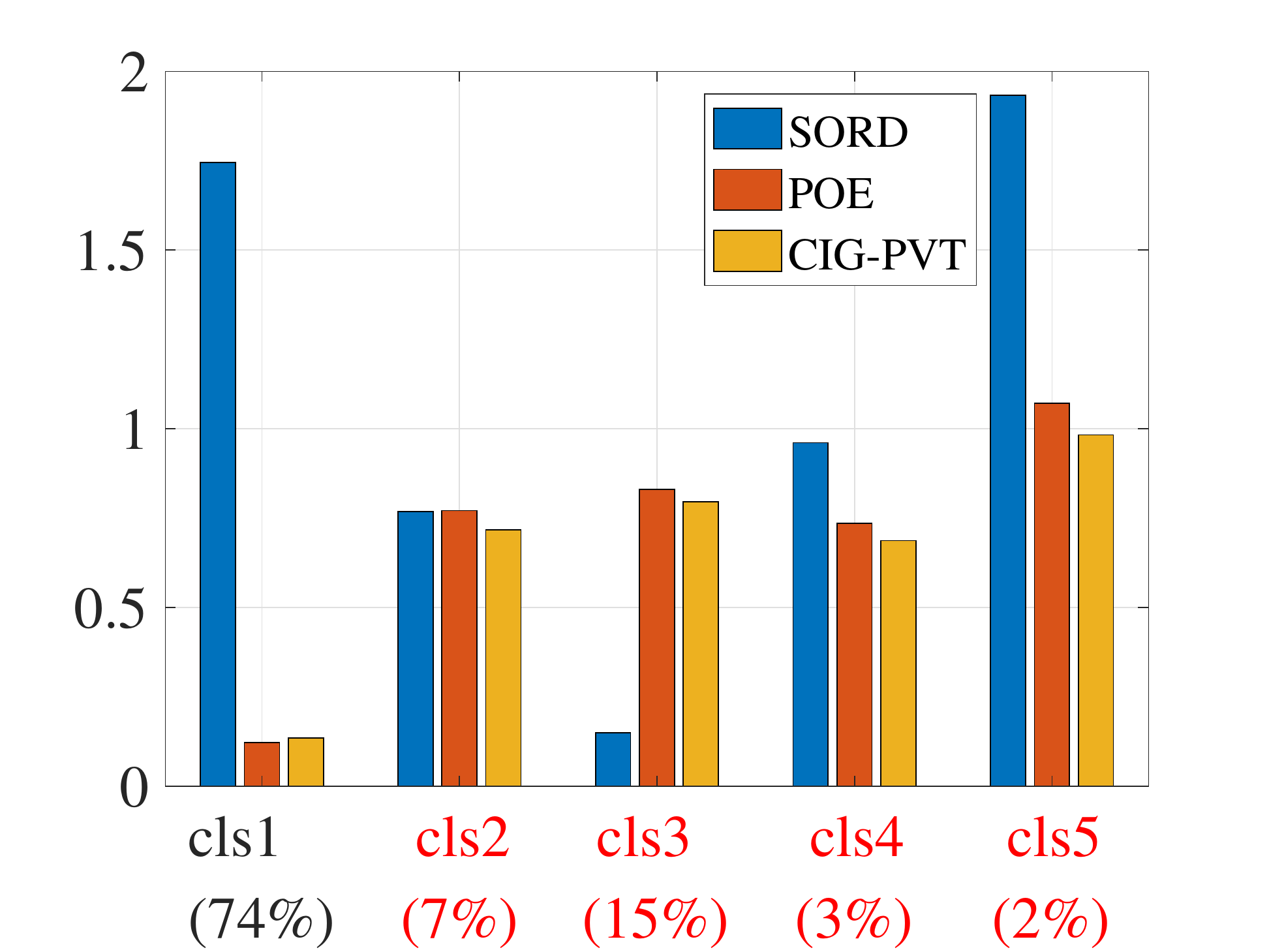}
         \caption{MAE on DR}
         \label{fig:robust_mae_dr}
     \end{subfigure}
     %\hfill
     \hspace{-3.75ex}
     \begin{subfigure}[b]{0.27\textwidth}
         \centering
         \includegraphics[width=\textwidth]{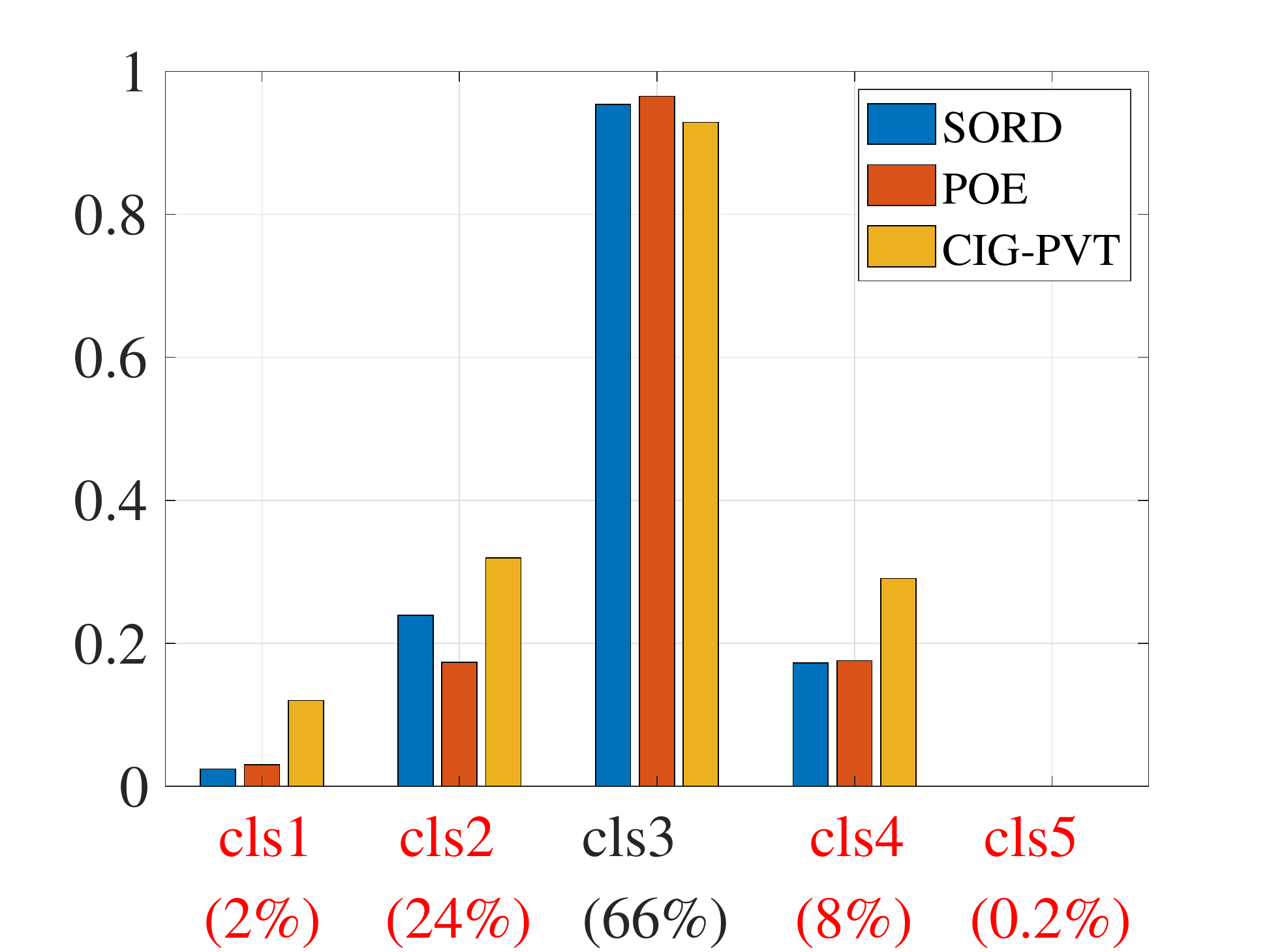}
         \caption{ACC on Aesthetics}
         \label{fig:robust_acc_aes}
     \end{subfigure}
     %\hfill 
    \hspace{-3.75ex}
     \begin{subfigure}[b]{0.27\textwidth}
         \centering
         \includegraphics[width=\textwidth]{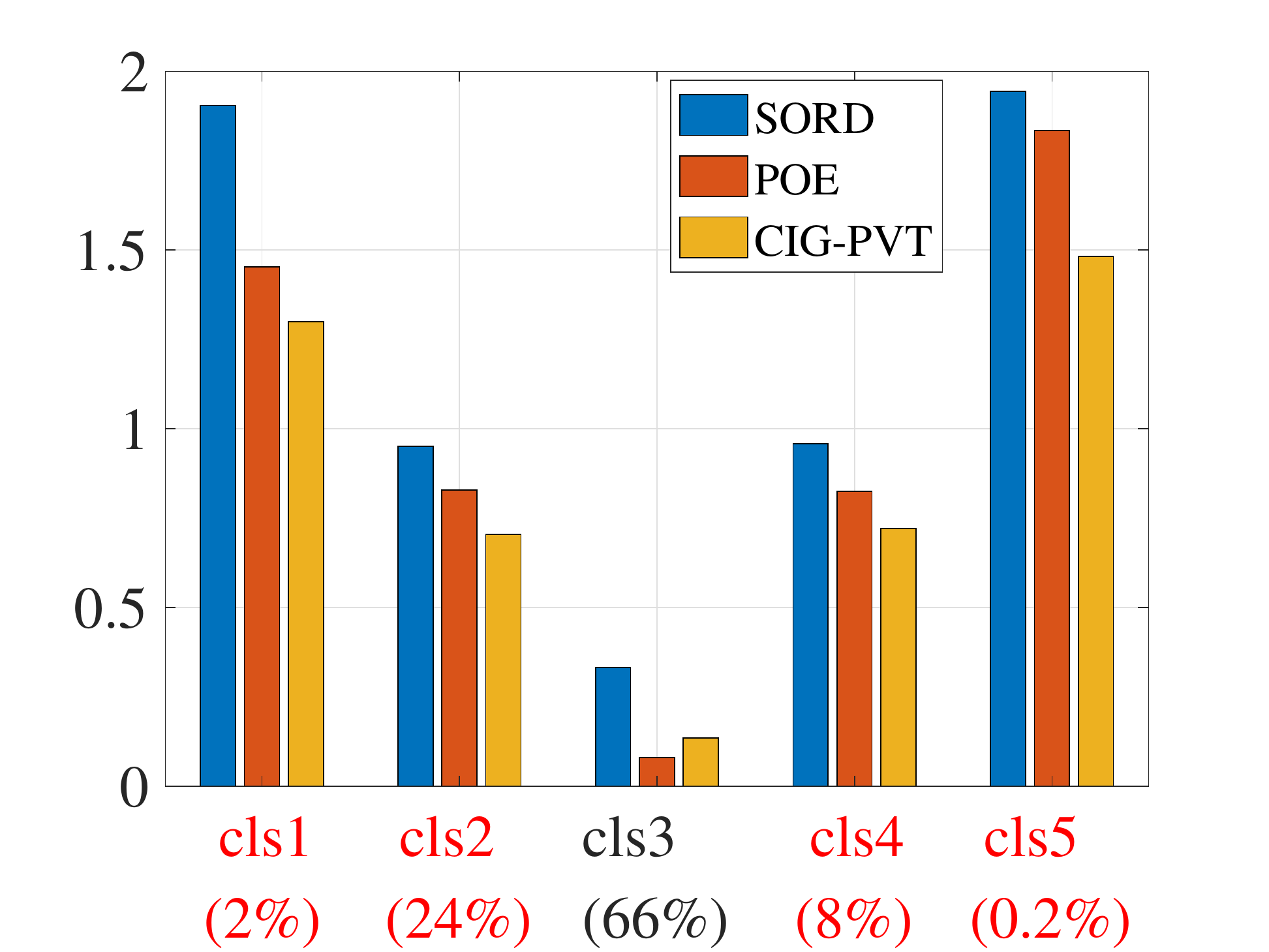}
         \caption{MAE on Aesthetics}
         \label{fig:robust_mae_aes}
     \end{subfigure}
    \caption{Detailed effectiveness for each category on DR and Aesthetics. Minority categories are marked in red. The results on Adience are given in the Supplementary Material.}
    \label{fig:robustness}
\end{figure*}

\subsection{Comparison with Known Methods} 
\label{subset:exp-overall}

We first evaluate the effectiveness of our approach by comparing with seven image ordinal regression methods (including state-of-the-art ones). 
The results are presented in Table~\ref{tab:previous compare}, in which \model-VGG and \model-PVT denote two variants of our approach based on the CNN and Transformer encoders, respectively. From the results, we observe the following.

Overall, the more recent SORD, POE, and MWR methods perform better than the other baselines. More specifically, MWR gives the best results among these three methods on Adience. This is because MWR further exploits the fine-grained categorical information (\eg concrete human ages) to refine the results, while such information is not available on the other two datasets. SORD attains competitive classification accuracy due to its soft label design. However, soft labels could introduce noise in ground-truth category probabilities, yielding large MAE by SORD, especially on DR and Aesthetics. On the other hand, POE shows good classification accuracy and MAE by aggregating results that correspond to different samples from an estimated distribution.

Our \model-PVT model using the Transformer encoder consistently outperforms all the baselines in both ACC and MAE on the three datasets. Indeed, our absolute improvements in ACC ($\uparrow$) and MAE ($\downarrow$) are (1.8\%, 0.4\%, 0.21\%) and (0.02, 0.008, 0.008) 
%\eat{0.007} 
on Adience, DR, and Aesthetics, respectively.
The improvement on Adience by \model is bigger.
This is probably because samples in Adience are distributed more evenly in categories and the categorical and structural information of face images can be well recognized and separated, enabling \model to generate more reliable fusion images.
On the other hand, on DR and Aesthetics, over 2/3 of the images belong to the same classes, thus making the overall improvements less significant after averaging on a large quantity of relatively `easy' images. But, such a class imbalance issue could severely affect the effectiveness on minority categories (Section~\ref{subset:exp-robust}). Our \model is designed to better classify those `hard' images for minority categories. In this sense, the smaller improvements of the overall ACC and MAE are still crucial.

\begin{table}[ht]
    \centering
    \begin{tabular}{cccc}
        \hline
        Dataset     & Method        & ACC (\%)$\uparrow$ & MAE$\downarrow$ \\
        \hline
        \multirow{3}{*}{Adience}     & SORD          & 32.75\eat{35.26} \eat{($\Delta=-5.1$)} & 1.533 \\
                   & POE           & \underline{36.26} \eat{($\Delta=-3.46$)} & \underline{0.749} \\
                   & \model-PVT          & \textbf{41.10\eat{40.36}} & \textbf{0.685} \\
        \hline
        \multirow{3}{*}{DR}          & SORD          & 23.78 \eat{($\Delta=-20.37$)} & \textbf{0.522} \\
                   & POE           & \underline{38.63} \eat{($\Delta=-5.52$)} & 0.824 \\
                   & \model-PVT          & \textbf{44.16} & \underline{0.794} \\
        \hline
        \multirow{3}{*}{Aesthetics}  & SORD          & 14.23\eat{($\Delta=-11.18$)} & 1.148 \\
                   & POE           & \underline{16.53} \eat{($\Delta=-11.46$)} & \underline{0.868} \\
                   & \model-PVT           & \textbf{25.41} & \textbf{0.840} \\
        \hline
    \end{tabular}
    \caption{Effectiveness comparison for minority categories.}
    \label{tab:minority}
\end{table}

\subsection{Robustness on Minority Categories} 
\label{subset:exp-robust}
Next, we examine the robustness of different methods for minority categories. 
Categories on which the classification accuracy is much lower than the highest one are taken as minorities, which are categories $\{4,6,7,8\}$, $\{2,3,4,5\}$, and $\{1,2,4,5\}$ in our three datasets, respectively. Based on the results in Table~\ref{tab:previous compare}, we only compare \model-PVT with SORD and POE in this set of tests. The overall and detailed results are presented in Table~\ref{tab:minority} and Fig.~\ref{fig:robustness}. We observe the following.

First, the performance on minority categories is worse for all the methods. 
For instance, the ACCs of (SORD, POE, and our \model-PVT) are (29\%, 20\%, 22\%), (54\%, 42\%, 39\%), and (56\%, 52\%, 45\%) lower than the overall accuracy on the three datasets, respectively.
These results empirically support our hypothesis that the class imbalance issue should be carefully addressed for image ordinal regression tasks. 

Second, our \model-PVT consistently outperforms the other two baselines on minority categories of all the datasets, except for MAE on DR.
Note that the soft labels of SORD are prone to decreasing the MAE for the `middle' categories, but in the cost of higher MAE for the leftmost and rightmost categories (see Figs.~\ref{fig:robust_mae_dr}\&\ref{fig:robust_mae_aes}). Yet, the lower MAE by SORD on cls3 of DR does not increase its corresponding ACC (Fig.~\ref{fig:robust_acc_dr}).

Third, the performance improvements of \model-PVT on minority categories are more substantial. Specifically, on DR, \model-PVT yields better ACC on two of the four minority categories and is on par with the best in one category (Fig.~\ref{fig:robust_acc_dr}). For the minority categories of Aesthetics, our results in ACC and MAE are consistently the best (Figs.~\ref{fig:robust_acc_aes}\&\ref{fig:robust_mae_aes}). Overall, the ACC is improved by (8.4\%, 4.8\%),
%\eat{1.7}\%), 
(20.4\%, 5.5\%), and (11.2\%, 8.9\%), and the MAE is decreased by (0.848, 0.064), ($-0.272$, 0.03), and (0.308, 0.028) compared to SORD and POE on the minority categories of the three datasets, respectively. 
These results verify that the controllable image generation process is effective in producing additional useful training samples for the less-represented categories to facilitate learning more accurate category boundaries, and our \model is more robust compared with the known methods.

\begin{table}[t]
    \centering
    %\begin{tabular}{m{1cm}<{\centering}m{1cm}<{\centering}m{1cm}<{\centering}m{1cm}<{\centering}m{1.5cm}<{\centering}m{1cm}<{\centering}}
    \begin{tabular}{c c c c c c}
        \hline
        Method      & IG               & S-F  &   CT  & ACC (\%)$\uparrow$               & MAE$\downarrow$ \\
        \hline
        \multirow{4}{*}{VGG}          & -                 & -                 & -                 & 57.4                  & 0.550  \\
                    & \checkmark        & -                 & -                 & 58.2                  & 0.532  \\
                    & \checkmark        & \checkmark        & -                 & \uline{61.0}          & \uline{0.485}  \\
                    & \checkmark        & \checkmark        & \checkmark        & \textbf{61.4}         & \textbf{0.471}  \\
        \hline
        \multirow{4}{*}{PVT}          & -                 & -                 & -                 & 61.6                  & 0.468  \\
                    & \checkmark        & -                 & -                 & 63.3                  & 0.458  \\
                    & \checkmark        & \checkmark        & -                 & \uline{63.9}          & \uline{0.447} \\
                    & \checkmark        & \checkmark        & \checkmark        & \textbf{64.4}         & \textbf{0.434} \\
        \hline
        \multirow{4}{*}{POE}          & -                 & -                 & -                 & 59.3                  & 0.485  \\
                    & \checkmark        & -                 & -                 &       60.5          &   0.475\\
                    & \checkmark        & \checkmark        & -                 & \uline{61.1}          & \uline{0.471}  \\
                    & \checkmark        & \checkmark        & \checkmark        & \textbf{61.6}         & \textbf{0.463}  \\
        \hline
    \end{tabular}
    \caption{Results of ablation study on Adience. 
    %\eat{IG, S-F, and CT stand for Image Generation, Separation-Fusion, and Continued Training.}
    }
    \label{tab:adience ablation}
\end{table}

\subsection{Ablation Study}
\label{subset:exp-ablation}

In the third set of tests, we conduct ablation study to empirically verify the rationality of our \model approach. We consider the following designs/variants of \model: direct image generation (IG) by simply adding the feature maps of the images, controllable image generation with the separation-fusion module (S-F), and continued training (CT) for the encoder. Moreover, we use CNN- and Transformer-based encoders (\ie VGG and PVT) as well as an existing image ordinal regression model POE as the backbone to investigate the applicability of \model.
Due to the page limit, we report only the results on Adience in Table~\ref{tab:adience ablation}.

We find that the fusion-based image generation strategy is effective to deal with the class imbalance issue in image ordinal regression. With IG, ACC and MAE of the three backbone models VGG, PVT, and POE are improved by (0.8\%, 1.7\%, 1.2\%) and (0.018, 0.010, 0.010), respectively. Moreover, the S-F module can further improve the performances, by (2.8\%, 0.6\%, 0.6\%) in ACC and (0.047, 0.011, 0.004) in MAE, indicating that our controllable image generation by separating the structural and categorical information is more reliable. Finally, our complete \model with continued training consistently yields the best results, \ie CT increases ACC by (0.4\%, 0.5\%, 0.5\%) and decreases MAE by (0.014, 0.013, 0.008). This is probably because the encoder could better focus on extracting category-related information of images in the continued training phase. 
Overall, we show that our designs of \model are generally useful, and together they assure the effectiveness of \model as a whole. In addition, one can see that \model is plug-and-play and flexible, and off-the-shelf image encoders or models can be readily integrated with \model to bring further improvement for image ordinal regression.

\begin{table}[t!]
    \centering
    % \begin{tabular}{m{1.4cm}<{\centering}m{1.7cm}<{\centering}|m{1.2cm}<{\centering}m{1.2cm}<{\centering}}
    \begin{tabular}{c|c|c}
        \hline
        Sampler   & ACC (\%)$\uparrow$ & MAE$\downarrow$ \\
        \hline 
        Equal &  60.73             & 0.472     \\
        Inverse-ratio & \textbf{61.38}    & \textbf{0.471}     \\
        \hline
        % Equal &  same       & 81.34             & 0.250     \\
        % Ratio &  same       & \textbf{81.56}    & \textbf{0.249}     \\
        % \hline
    \end{tabular}
    \caption{The impacts of two samplers on Adience.}
    \label{tab:sampler}
\end{table}

We also test the effectiveness of two different samplers (\ie equal and inverse-ratio) for \model. The results on Adience are reported in Table~\ref{tab:sampler}. We find that the inverse-ratio sampler is better since it can supplement more images for minority categories than the equal sampler.

\subsection{Parameter Sensitivity}
\label{subset:exp-para}

\begin{figure}[t!]
    \centering
    \includegraphics[scale=0.3]{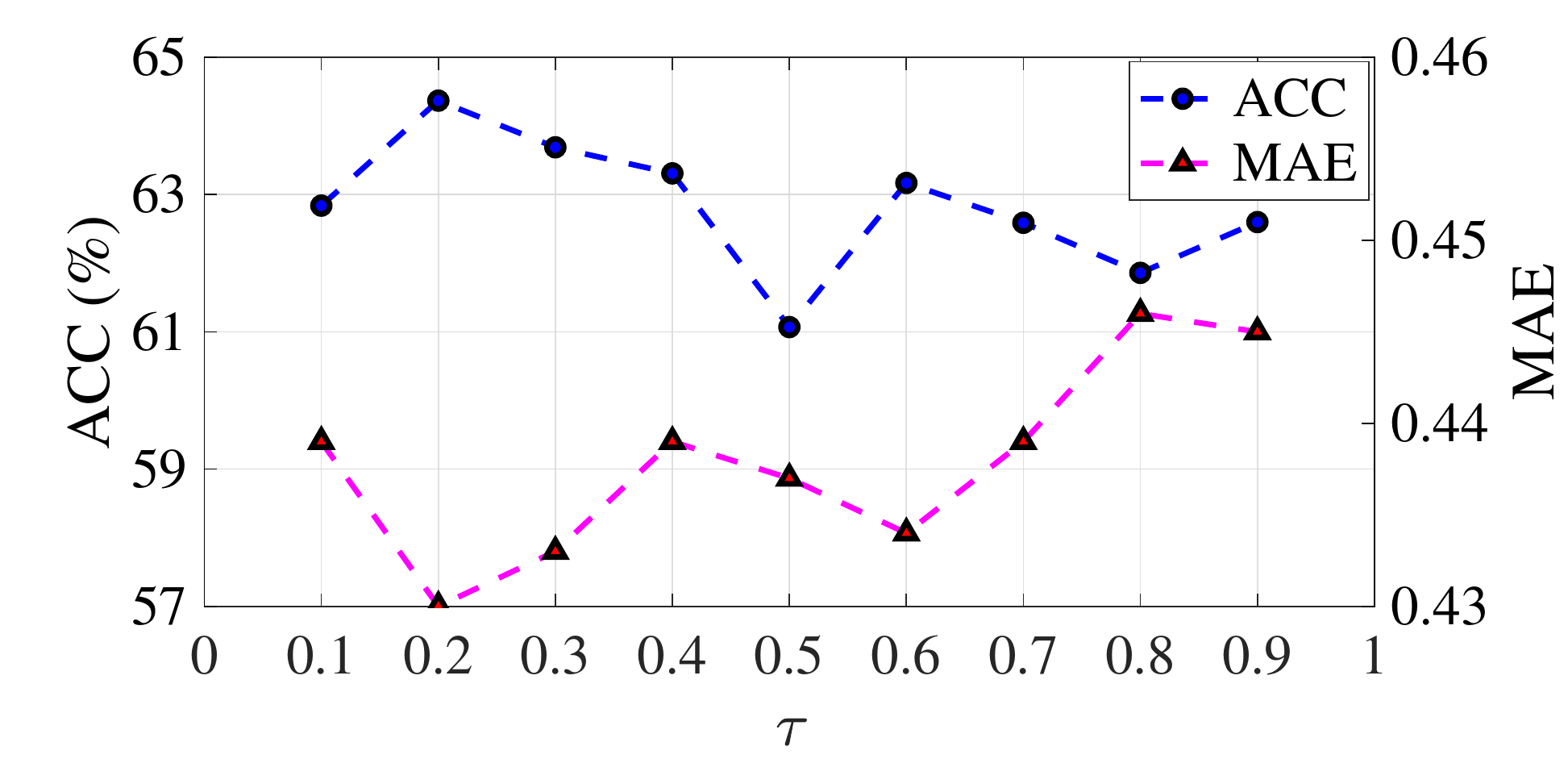}
    \caption{The impact of $\tau$ on Adience.} % Proportion of structural and categorical components in S-F: the proportion of structural information is $\tau$ and that for categorical information is $1-\tau$.
    \label{fig:ratio}
\end{figure}

Finally, we evaluate the parameter sensitivity of \model. 
To examine the impact of $\tau$, \ie the length parameter for the structural and categorical information vectors in the separation-fusion module, we vary $\tau$ from 0.1 to 0.9, fix the other parameters to their default values, and test the ACC and MAE results, as shown in Fig.~\ref{fig:ratio}.
When increasing $\tau$, ACC first increases and then decreases in general with the increment of $\tau$, and MAE first decreases and then increases on the contrary. The best ACC and MAE are attained at $\tau=0.2$. This implies that the image encoder uses more bits (in the categorical information vectors) to maintain categorical information of images.
The results for the other hyper-parameters (\ie $\lambda$, $\alpha$, and $\beta$) are given in the Supplementary Material.

% this phenomenon is consistent with our conclusion. The purpose of the image encoder is for classification, under this premise, the feature map extracted by the encoder contains more categorical information. On the other hand, the category feature determines the controllable generation process, because this feature guides the category of the generated image.

% \begin{figure}[t!]
%     \centering
%     \includegraphics[scale=0.2]{img/ratio_subplot.eps}
%     \caption{Ratio in SF: the proportion of structural information is $Ratio$ and that for categorical information is $1-Ratio$.}
%     \label{fig:ratio}
% \end{figure}

\section{Conclusions}

In this paper, we focused on the class imbalance and category overlap issues in image ordinal regression which have been largely overlooked. We proposed, to our best knowledge, the first image ordinal regression approach that
%in the literature 
directly addresses these two issues. 
We presented a novel framework \model based on controllable image generation which can generate artificial images to facilitate learning more accurate and robust decision boundaries. Each generated image contains structural information of one image and categorical information of another image from adjacent categories. To achieve such controllable generation, \model was designed to learn the separation of the structural and categorical information of images using three self-supervised objectives.
Extensive experiments on three different image ordinal regression scenarios verified the effectiveness and robustness of \model compared with state-of-the-art methods. More specifically, \model based on the Transformer encoder established new best-known performances. We also empirically showed that previous methods incurred considerable robustness issues on minority categories, and our \model approach yielded higher improvements on such categories.
It is expected that our work will inspire further studies on robust image ordinal regression.

% \section*{Ethical Statement}

% There are no ethical issues.

\section*{Acknowledgments}

This research was partially supported by National Key R\&D Program of China under grant No. 2018AAA0102100, National Natural Science Foundation of China under grants No. 62106218 and No. 62176231.

\bibliographystyle{named}
% \bibliography{ijcai23}
% \bibliography{ijcai23_reference}

\begin{thebibliography}{}

\bibitem[\protect\citeauthoryear{Beckham and Pal}{2017}]{Poisson}
Christopher Beckham and Christopher~J. Pal.
\newblock Unimodal probability distributions for deep ordinal classification.
\newblock In Doina Precup and Yee~Whye Teh, editors, {\em 34th International
  Conference on Machine Learning}, volume~70 of {\em Proceedings of Machine
  Learning Research}, pages 411--419. {PMLR}, 2017.

\bibitem[\protect\citeauthoryear{Bulten \bgroup \em et al.\egroup
  }{2022}]{prostate}
Wouter Bulten, Kimmo Kartasalo, Po-Hsuan~Cameron Chen, Peter Str{\"o}m, Hans
  Pinckaers, Kunal Nagpal, Yuannan Cai, David~F Steiner, Hester van Boven,
  Robert Vink, et~al.
\newblock Artificial intelligence for diagnosis and {Gleason} grading of
  prostate cancer: The {PANDA} challenge.
\newblock {\em Nature Medicine}, 28(1):154--163, 2022.

\bibitem[\protect\citeauthoryear{Chen \bgroup \em et al.\egroup
  }{2022}]{blasto}
Tingting Chen, Yi~Cheng, Jinhong Wang, Zhaoxia Yang, Wenhao Zheng, Danny~Z.
  Chen, and Jian Wu.
\newblock Automating blastocyst formation and quality prediction in time-lapse
  imaging with adaptive key frame selection.
\newblock In Linwei Wang, Qi~Dou, P.~Thomas Fletcher, Stefanie Speidel, and
  Shuo Li, editors, {\em Medical Image Computing and Computer Assisted
  Intervention, Part {IV}}, volume 13434 of {\em Lecture Notes in Computer
  Science}, pages 445--455. Springer, 2022.

\bibitem[\protect\citeauthoryear{Decenci{\`e}re \bgroup \em et al.\egroup
  }{2014}]{Messidor}
Etienne Decenci{\`e}re, Xiwei Zhang, Guy Cazuguel, Bruno Lay, B{\'e}atrice
  Cochener, Caroline Trone, Philippe Gain, Richard Ordonez, Pascale Massin, Ali
  Erginay, et~al.
\newblock Feedback on a publicly distributed image database: The {Messidor}
  database.
\newblock {\em Image Analysis \& Stereology}, 33(3):231--234, 2014.

\bibitem[\protect\citeauthoryear{Deng \bgroup \em et al.\egroup
  }{2009}]{imagenet}
Jia Deng, Wei Dong, Richard Socher, Li{-}Jia Li, Kai Li, and Li~Fei{-}Fei.
\newblock {ImageNet}: {A} large-scale hierarchical image database.
\newblock In {\em 2009 {IEEE} Computer Society Conference on Computer Vision
  and Pattern Recognition}, pages 248--255. {IEEE} Computer Society, 2009.

\bibitem[\protect\citeauthoryear{Diaz and Marathe}{2019}]{sord}
Raul Diaz and Amit Marathe.
\newblock Soft labels for ordinal regression.
\newblock In {\em {IEEE} Conference on Computer Vision and Pattern
  Recognition}, pages 4738--4747. Computer Vision Foundation / {IEEE}, 2019.

\bibitem[\protect\citeauthoryear{Dosovitskiy \bgroup \em et al.\egroup
  }{2021}]{vit}
Alexey Dosovitskiy, Lucas Beyer, Alexander Kolesnikov, Dirk Weissenborn,
  Xiaohua Zhai, Thomas Unterthiner, Mostafa Dehghani, Matthias Minderer, Georg
  Heigold, Sylvain Gelly, Jakob Uszkoreit, and Neil Houlsby.
\newblock An image is worth 16x16 words: Transformers for image recognition at
  scale.
\newblock In {\em 9th International Conference on Learning Representations}.
  OpenReview.net, 2021.

\bibitem[\protect\citeauthoryear{Fu and Huang}{2008}]{regression_face}
Yun Fu and Thomas~S. Huang.
\newblock Human age estimation with regression on discriminative aging
  manifold.
\newblock {\em {IEEE} Trans. Multim.}, 10(4):578--584, 2008.

\bibitem[\protect\citeauthoryear{Fu \bgroup \em et al.\egroup
  }{2018}]{ranking2}
Huan Fu, Mingming Gong, Chaohui Wang, Kayhan Batmanghelich, and Dacheng Tao.
\newblock Deep ordinal regression network for monocular depth estimation.
\newblock In {\em 2018 {IEEE} Conference on Computer Vision and Pattern
  Recognition}, pages 2002--2011. Computer Vision Foundation / {IEEE} Computer
  Society, 2018.

\bibitem[\protect\citeauthoryear{Geiger \bgroup \em et al.\egroup
  }{2013}]{KITTI}
Andreas Geiger, Philip Lenz, Christoph Stiller, and Raquel Urtasun.
\newblock Vision meets robotics: The {KITTI} dataset.
\newblock {\em Int. J. Robotics Res.}, 32(11):1231--1237, 2013.

\bibitem[\protect\citeauthoryear{Guo and Mu}{2013a}]{DBLP:conf/fgr/GuoM13}
Guodong Guo and Guowang Mu.
\newblock Joint estimation of age, gender and ethnicity: {CCA} vs. {PLS}.
\newblock In {\em 10th {IEEE} International Conference and Workshops on
  Automatic Face and Gesture Recognition}, pages 1--6. {IEEE} Computer Society,
  2013.

\bibitem[\protect\citeauthoryear{Guo and Mu}{2013b}]{regression_face2}
Guodong Guo and Guowang Mu.
\newblock Joint estimation of age, gender and ethnicity: {CCA} vs. {PLS}.
\newblock In {\em 10th {IEEE} International Conference and Workshops on
  Automatic Face and Gesture Recognition}, pages 1--6. {IEEE} Computer Society,
  2013.

\bibitem[\protect\citeauthoryear{He \bgroup \em et al.\egroup }{2022}]{mae}
Kaiming He, Xinlei Chen, Saining Xie, Yanghao Li, Piotr Doll{\'{a}}r, and
  Ross~B. Girshick.
\newblock Masked autoencoders are scalable vision learners.
\newblock In {\em {IEEE/CVF} Conference on Computer Vision and Pattern
  Recognition}, pages 15979--15988. {IEEE}, 2022.

\bibitem[\protect\citeauthoryear{Levi and Hassner}{2015}]{adience}
Gil Levi and Tal Hassner.
\newblock Age and gender classification using convolutional neural networks.
\newblock In {\em 2015 {IEEE} Conference on Computer Vision and Pattern
  Recognition Workshops}, pages 34--42. {IEEE} Computer Society, 2015.

\bibitem[\protect\citeauthoryear{Li \bgroup \em et al.\egroup }{2021}]{poe}
Wanhua Li, Xiaoke Huang, Jiwen Lu, Jianjiang Feng, and Jie Zhou.
\newblock Learning probabilistic ordinal embeddings for uncertainty-aware
  regression.
\newblock In {\em {IEEE} Conference on Computer Vision and Pattern
  Recognition}, pages 13896--13905. Computer Vision Foundation / {IEEE}, 2021.

\bibitem[\protect\citeauthoryear{Liu \bgroup \em et al.\egroup }{2018a}]{dr}
Xiaofeng Liu, Yang Zou, Yuhang Song, Chao Yang, Jane You, and B.~V. K.~Vijaya
  Kumar.
\newblock Ordinal regression with neuron stick-breaking for medical diagnosis.
\newblock In Laura Leal{-}Taix{\'{e}} and Stefan Roth, editors, {\em {ECCV}
  Workshops, Proceedings, Part {VI}}, volume 11134 of {\em Lecture Notes in
  Computer Science}, pages 335--344. Springer, 2018.

\bibitem[\protect\citeauthoryear{Liu \bgroup \em et al.\egroup
  }{2018b}]{cnnpor}
Yanzhu Liu, Adams~Wai{-}Kin Kong, and Chi~Keong Goh.
\newblock A constrained deep neural network for ordinal regression.
\newblock In {\em 2018 {IEEE} Conference on Computer Vision and Pattern
  Recognition}, pages 831--839. Computer Vision Foundation / {IEEE} Computer
  Society, 2018.

\bibitem[\protect\citeauthoryear{Liu \bgroup \em et al.\egroup
  }{2019}]{gpdnnor}
Yanzhu Liu, Fan Wang, and Adams~Wai{-}Kin Kong.
\newblock Probabilistic deep ordinal regression based on {Gaussian} processes.
\newblock In {\em 2019 {IEEE/CVF} International Conference on Computer Vision},
  pages 5300--5308. {IEEE}, 2019.

\bibitem[\protect\citeauthoryear{Liu \bgroup \em et al.\egroup
  }{2021}]{orthogonal}
Lina Liu, Xibin Song, Mengmeng Wang, Yong Liu, and Liangjun Zhang.
\newblock Self-supervised monocular depth estimation for all day images using
  domain separation.
\newblock In {\em 2021 {IEEE/CVF} International Conference on Computer Vision},
  pages 12717--12726. {IEEE}, 2021.

\bibitem[\protect\citeauthoryear{Lukyanenko \bgroup \em et al.\egroup
  }{2021}]{embryo}
Stanislav Lukyanenko, Won{-}Dong Jang, Donglai Wei, Robbert Struyven, Yoon Kim,
  Brian~D. Leahy, Helen~Y. Yang, Alexander~M. Rush, Dalit Ben{-}Yosef, Daniel
  Needleman, and Hanspeter Pfister.
\newblock Developmental stage classification of embryos using two-stream neural
  network with linear-chain conditional random field.
\newblock In Marleen de~Bruijne, Philippe~C. Cattin, St{\'{e}}phane Cotin,
  Nicolas Padoy, Stefanie Speidel, Yefeng Zheng, and Caroline Essert, editors,
  {\em Medical Image Computing and Computer Assisted Intervention, Part
  {VIII}}, volume 12908 of {\em Lecture Notes in Computer Science}, pages
  363--372. Springer, 2021.

\bibitem[\protect\citeauthoryear{Niu \bgroup \em et al.\egroup
  }{2016}]{niu2016ordinal}
Zhenxing Niu, Mo~Zhou, Le~Wang, Xinbo Gao, and Gang Hua.
\newblock Ordinal regression with multiple output {CNN} for age estimation.
\newblock In {\em Proceedings of the IEEE Conference on Computer Vision and
  Pattern Recognition}, pages 4920--4928, 2016.

\bibitem[\protect\citeauthoryear{Palermo \bgroup \em et al.\egroup
  }{2012}]{Historical}
Frank Palermo, James Hays, and Alexei~A. Efros.
\newblock Dating historical color images.
\newblock In Andrew~W. Fitzgibbon, Svetlana Lazebnik, Pietro Perona, Yoichi
  Sato, and Cordelia Schmid, editors, {\em 12th European Conference on Computer
  Vision, Part {VI}}, volume 7577 of {\em Lecture Notes in Computer Science},
  pages 499--512. Springer, 2012.

\bibitem[\protect\citeauthoryear{Paszke \bgroup \em et al.\egroup
  }{2019}]{pytorch}
Adam Paszke, Sam Gross, Francisco Massa, Adam Lerer, James Bradbury, Gregory
  Chanan, Trevor Killeen, Zeming Lin, Natalia Gimelshein, Luca Antiga, Alban
  Desmaison, Andreas Kopf, Edward Yang, Zachary DeVito, Martin Raison, Alykhan
  Tejani, Sasank Chilamkurthy, Benoit Steiner, Lu~Fang, Junjie Bai, and Soumith
  Chintala.
\newblock {PyTorch}: An imperative style, high-performance deep learning
  library.
\newblock In H.~Wallach, H.~Larochelle, A.~Beygelzimer, F.~d'Alché Buc,
  E.~Fox, and R.~Garnett, editors, {\em Advances in Neural Information
  Processing Systems}, pages 8024--8035. Curran Associates, Inc., 2019.

\bibitem[\protect\citeauthoryear{Ratner \bgroup \em et al.\egroup }{2018}]{MT}
Vadim Ratner, Yoel Shoshan, and Tal Kachman.
\newblock Learning multiple non-mutually-exclusive tasks for improved
  classification of inherently ordered labels.
\newblock {\em CoRR}, abs/1805.11837, 2018.

\bibitem[\protect\citeauthoryear{Ronneberger \bgroup \em et al.\egroup
  }{2015}]{unet}
Olaf Ronneberger, Philipp Fischer, and Thomas Brox.
\newblock {U-Net}: Convolutional networks for biomedical image segmentation.
\newblock In Nassir Navab, Joachim Hornegger, William M.~Wells III, and
  Alejandro~F. Frangi, editors, {\em Medical Image Computing and
  Computer-Assisted Intervention, Part {III}}, volume 9351 of {\em Lecture
  Notes in Computer Science}, pages 234--241. Springer, 2015.

\bibitem[\protect\citeauthoryear{Schifanella \bgroup \em et al.\egroup
  }{2015}]{aesthetics}
Rossano Schifanella, Miriam Redi, and Luca~Maria Aiello.
\newblock An image is worth more than a thousand favorites: Surfacing the
  hidden beauty of {Flickr} pictures.
\newblock In Meeyoung Cha, Cecilia Mascolo, and Christian Sandvig, editors,
  {\em 9th International Conference on Web and Social Media}, pages 397--406.
  {AAAI} Press, 2015.

\bibitem[\protect\citeauthoryear{Shin \bgroup \em et al.\egroup }{2022}]{mwr}
Nyeong{-}Ho Shin, Seon{-}Ho Lee, and Chang{-}Su Kim.
\newblock Moving window regression: {A} novel approach to ordinal regression.
\newblock In {\em {IEEE/CVF} Conference on Computer Vision and Pattern
  Recognition}, pages 18739--18748. {IEEE}, 2022.

\bibitem[\protect\citeauthoryear{Simonyan and Zisserman}{2015}]{vgg}
Karen Simonyan and Andrew Zisserman.
\newblock Very deep convolutional networks for large-scale image recognition.
\newblock In Yoshua Bengio and Yann LeCun, editors, {\em 3rd International
  Conference on Learning Representations}, 2015.

\bibitem[\protect\citeauthoryear{Wang \bgroup \em et al.\egroup }{2021}]{pvt}
Wenhai Wang, Enze Xie, Xiang Li, Deng{-}Ping Fan, Kaitao Song, Ding Liang, Tong
  Lu, Ping Luo, and Ling Shao.
\newblock Pyramid vision {Transformer}: {A} versatile backbone for dense
  prediction without convolutions.
\newblock In {\em 2021 {IEEE/CVF} International Conference on Computer Vision},
  pages 548--558. {IEEE}, 2021.

\bibitem[\protect\citeauthoryear{Workman \bgroup \em et al.\egroup
  }{2016}]{HLW}
Scott Workman, Menghua Zhai, and Nathan Jacobs.
\newblock Horizon lines in the wild.
\newblock In Richard~C. Wilson, Edwin~R. Hancock, and William A.~P. Smith,
  editors, {\em Proceedings of the British Machine Vision Conference}. {BMVA}
  Press, 2016.

\bibitem[\protect\citeauthoryear{Zhu \bgroup \em et al.\egroup
  }{2017}]{cycleGAN}
Jun{-}Yan Zhu, Taesung Park, Phillip Isola, and Alexei~A. Efros.
\newblock Unpaired image-to-image translation using cycle-consistent
  adversarial networks.
\newblock In {\em {IEEE} International Conference on Computer Vision}, pages
  2242--2251. {IEEE} Computer Society, 2017.

\end{thebibliography}

\clearpage 
\section*{Appendix A: Robustness on Minority Categories on Adience}

\begin{figure}[ht]
    \centering
    \begin{subfigure}[b]{0.52\textwidth}
         \centering
         \hspace{-8ex}
         \includegraphics[width=\textwidth]{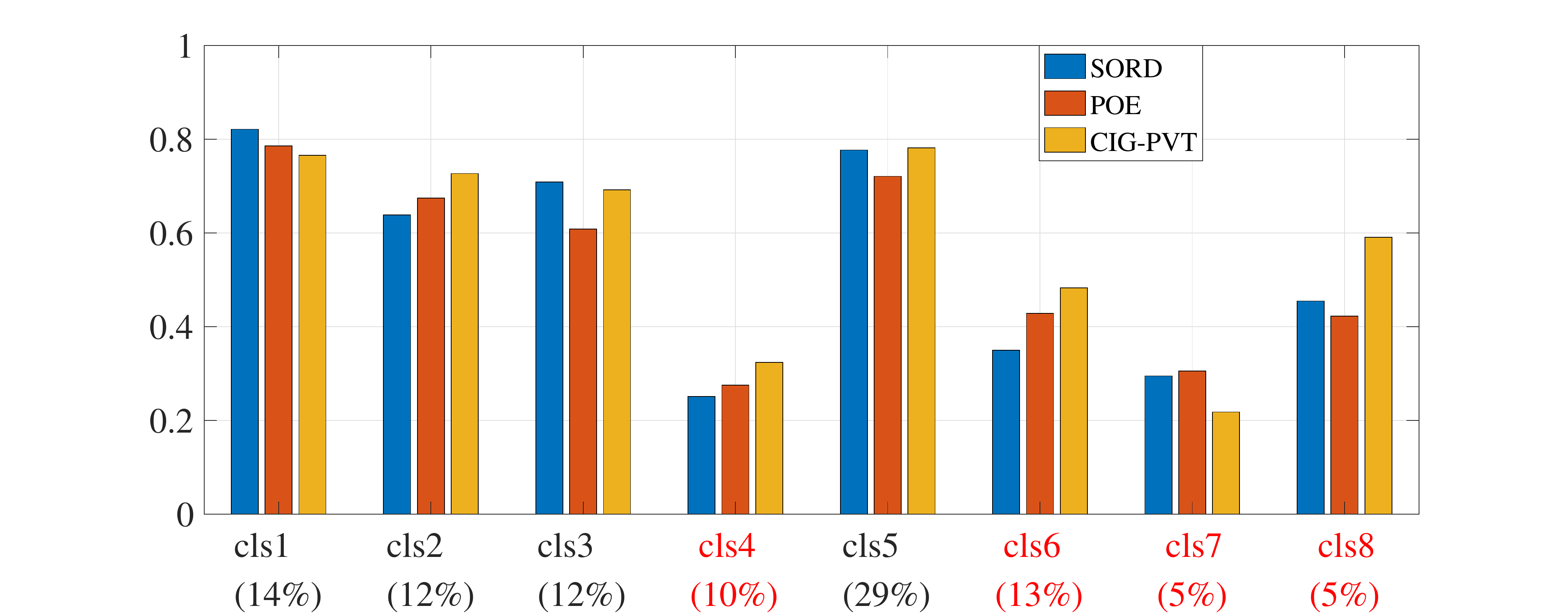}
         \caption{ACC}
         \label{fig:robust_acc_adience}
     \end{subfigure}
     \begin{subfigure}[b]{0.52\textwidth}
         \centering
         \hspace{-8ex}
         \includegraphics[width=\textwidth]{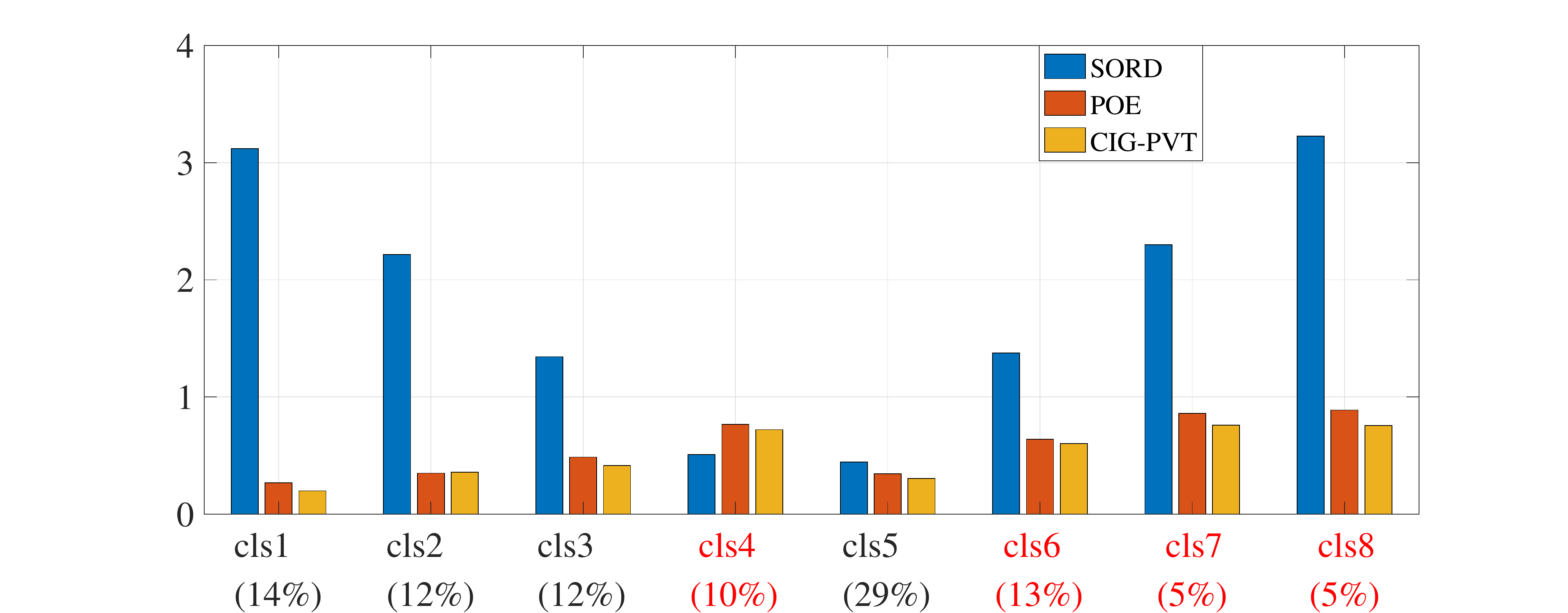}
         \caption{MAE}
         \label{fig:robust_mae_adience}
     \end{subfigure}
    \caption{Detailed effectiveness for each category on Adience. Minority categories are marked in red.}
    \label{fig:robustness_adience}
\end{figure}

Figure~\ref{fig:robustness_adience} presents the detailed effectiveness of SORD, POE, and \model-PVT for each category on the Adience dataset. 
Recall that our \model-PVT approach increases the ACC on minority categories by (8.4\%, 4.8\%) on this dataset, compared with (SORD, POE). Indeed, \model-PVT obtains the best ACC on three of the four minority categories, as shown in Fig.~\ref{fig:robust_acc_adience}. It performs worse on cls7, which is possibly a trade-off for the improved performance on the adjacent categories. Similarly, \model-PVT decreases the MAE by (0.848, 0.064) compared with (SORD, POE). From Fig.~\ref{fig:robust_mae_adience} we can observe that \model-PVT could effectively optimize the MAE on almost all categories, except for the `middle' cls4 on which SORD is partial.
The above results lead to the same conclusion that \model-PVT has better robustness for minority categories.

\section*{Appendix B: Extra Results for Parameter Sensitivity}

\begin{figure}[h]
    \centering
    \includegraphics[scale=0.32]{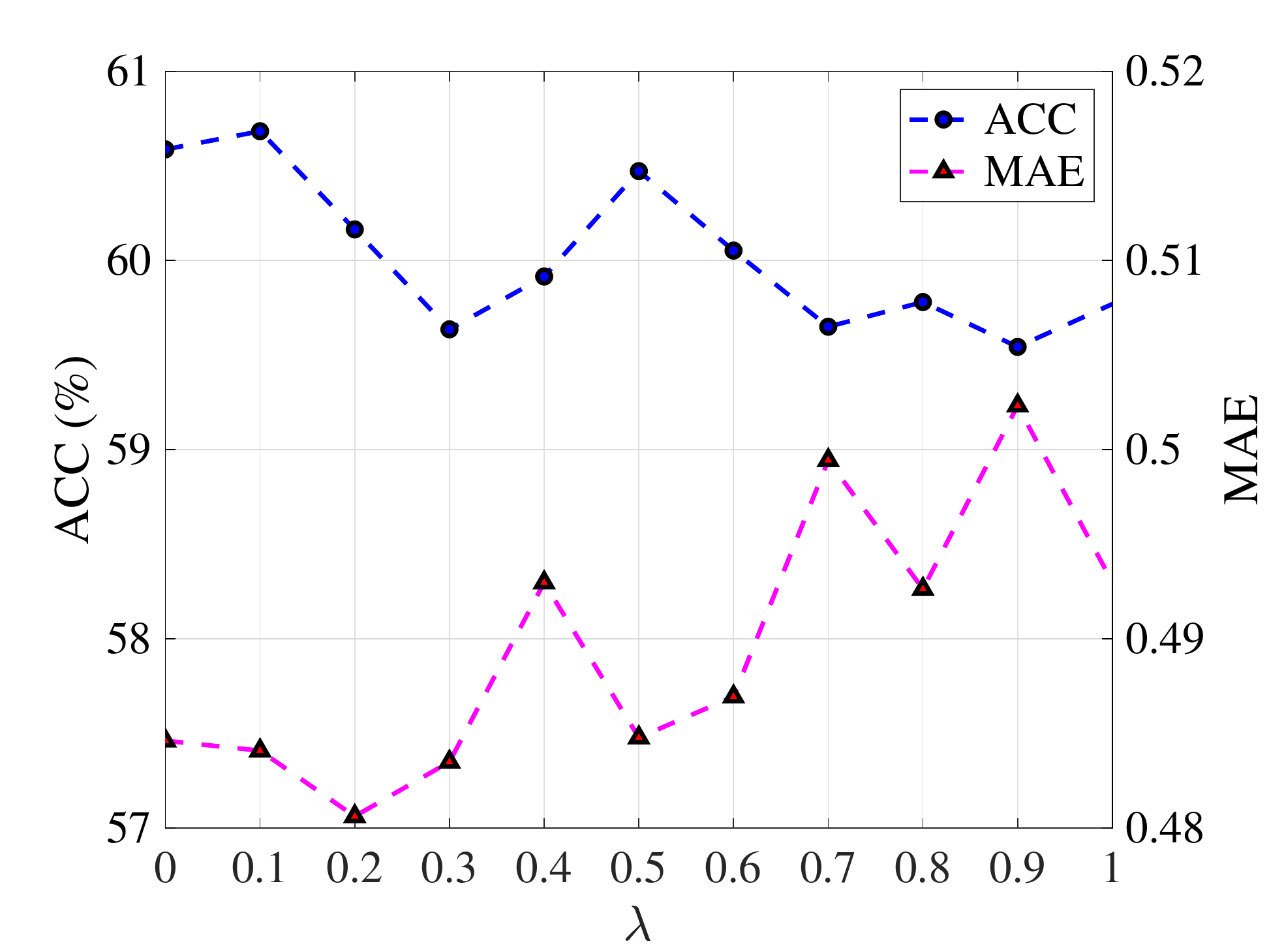}
    \caption{Impacts of $\lambda$ on Adience} % Plots of ACC and MAE with respect to the values of hyperparameter $\lambda$ using five-cross validation on the Adience dataset.
    \label{lambda}
\end{figure}

We further present the sensitivity results of hyper-parameter $\lambda$, $\alpha$, and $\beta$. 
The hyper-parameter $\lambda$ in Eq.~(7) determines the influence of fusion images in the classification loss. To evaluate the impacts of $\lambda$, we vary $\lambda$ from 0 to 1 with a step size of 0.1 and fix $\alpha$ and $\beta$ to 1, \ie $\lambda$ is the first optimized parameter.
The results on Aidence are shown in Fig.~\ref{lambda}, from which we find that the MAE decreases from 0 to 0.2 and then increases in general. Moreover, the ACC reaches its peak at $\lambda=0.1$ and decreases afterward. 
Overall, small values are preferred for $\lambda$. 
This is because in the early steps, the generator has not been well trained yet and the fusion images are not very reliable to be used for training. Setting $\lambda$ to relatively small values could prevent early pollution from the early-generated fusion images. In image ordinal regression tasks, MAE is regarded as more meaningful than ACC in general. Therefore, we choose $\lambda=0.2$ in our experiments by default.

%We first set the two hyper-parameters $\alpha$ and $\beta$ to 1 and use an add operation to replace the S-F module, and then optimize $\lambda$ from 0 to 1 with a step size of 0.1. 

%\section*{Appendix B: Experimental Results on $\alpha$ and $\beta$}

The generation loss in Eq.~(5) consists of a structural similarity term, a categorical similarity term, and a reconstruction loss. 
We use two hyper-parameters $\alpha$ and $\beta$ to regularize their contributions. 
To evaluate the impacts of $\alpha$ and $\beta$, we choose their values from $\{1,2,5\}$, fix $\lambda=0.2$, and test the performance. The results on Adience are reported in Table~\ref{tab:alpha-beta}.
We find that the combination of $\alpha=5$ and $\beta=1$ yields the highest ACC while the one with $\alpha=5$ and $\beta=2$ gives the lowest MAE. 
%Since the ACC difference between these two combinations is 0.28\% while the MAE difference is 0.007, we think the performance gap in MAE is larger and more important than ACC. 
We again choose default parameter values based on the MAE results, \ie setting $\alpha=5$ and $\beta=2$ in our experiments.
%Thus, $[5,2]$ is chosen for $\alpha$ and $\beta$. 
This result is also in line with our experience. In feature extraction, structural information is generally easier to extract, and the corresponding loss is smaller. We thus give the structural loss a greater weight to balance the contributions of the three terms.

\begin{table}[t]
    \centering
    \begin{tabular}{c|c|cc}
        \hline
        $\alpha$ & $\beta$      & ACC (\%)$\uparrow$               & MAE$\downarrow$ \\
        \hline 
        1&1         & 60.16                 & 0.481      \\
        1&2         & 60.11                 & 0.478     \\
        1&5         & 60.44                 & 0.484     \\
        2&1         & 60.74                 & 0.477     \\
        2&2         & 60.62                 & 0.490     \\
        2&5         & 59.45                 & 0.485     \\
        5&1         & \textbf{60.86}        & 0.474     \\
        5&2         & 60.58                 & \textbf{0.467} \\
        5&5         & 60.08                 & 0.495     \\
        \hline
    \end{tabular}
    \caption{Impaces of $\alpha$ and $\beta$ on Adience.}
    \label{tab:alpha-beta}
\end{table}

% \begin{table}[t]
%     \centering
%     \begin{tabular}{m{1.4cm}<{\centering}|m{1.4cm}<{\centering}m{3.5cm}<{\centering}}
%         \hline
%         Dataset     & Method        & Minority Class MAE\\
%         \hline
%         \multirow{3}{*}{Adience}     & SORD          & 1.506 ($\Delta=-0.802$)\\
%                    & POE           & 0.734 ($\Delta=-0.3$) \\
%                    & OURS          & \textbf{0.704}  \\
%         \hline
%         \multirow{3}{*}{DR}          & SORD          & \textbf{0.522} ($\Delta=0.273$)\\
%                    & POE           & 0.824 ($\Delta=-0.029$) \\
%                    & OURS          & 0.794  \\
%         \hline
%         \multirow{3}{*}{Aesthetics}  & SORD          & 1.148 ($\Delta=-0.308 $)\\
%                    & POE           & 0.938 ($\Delta=-0.098 $) \\
%                    & OURS          & \textbf{0.840}  \\
%         \hline
%     \end{tabular}
%     \caption{Minority MAE}
%     \label{tab:minority2}
% \end{table}

% \section*{Appendix D: MAE on Each Category}

\end{document}